\documentclass{article}

\usepackage{arxiv}

\usepackage[utf8]{inputenc} % allow utf-8 input
\usepackage[T1]{fontenc}    % use 8-bit T1 fonts
\usepackage{hyperref}       % hyperlinks
\usepackage{url}            % simple URL typesetting
\usepackage{booktabs}       % professional-quality tables
\usepackage{amsfonts}       % blackboard math symbols
\usepackage{nicefrac}       % compact symbols for 1/2, etc.
\usepackage{microtype}      % microtypography
\usepackage{lipsum}		% Can be removed after putting your text content
\usepackage{graphicx}
\usepackage{natbib}
\usepackage{doi}

\usepackage{graphicx} % DO NOT CHANGE THIS
\usepackage{natbib}  % DO NOT CHANGE THIS AND DO NOT ADD ANY OPTIONS TO IT
\usepackage{caption} % DO NOT CHANGE THIS AND DO NOT ADD ANY OPTIONS TO IT
\usepackage{booktabs}
\usepackage{multirow}
\usepackage{makecell}
\usepackage{adjustbox}
\usepackage[table]{xcolor}

\usepackage{algorithm}
\usepackage{algorithmic}
\usepackage{fontawesome}
\usepackage{amsmath}
\usepackage{amssymb}
\usepackage{subcaption}
\usepackage[dvipsnames]{xcolor}
\usepackage{newfloat}
\usepackage{listings}
\DeclareCaptionStyle{ruled}{labelfont=normalfont,labelsep=colon,strut=off} % DO NOT CHANGE THIS
\floatstyle{ruled}
\newfloat{listing}{tb}{lst}{}
\floatname{listing}{Listing}

\usepackage{booktabs}   % \toprule, \midrule, \bottomrule, \cmidrule
\usepackage{multirow}   % \multirow
\usepackage{makecell}   % \makecell
\usepackage{adjustbox}  % \begin{adjustbox}
\usepackage{xcolor}     % \textcolor
\usepackage{pifont}     % \ding
\usepackage{pifont}
\usepackage{placeins}
\newcommand{\cmark}{\ding{51}}
\title{Template-Search Domain Adaptation via Multi-Stage Feature Alignment for Cross-Modal Object Tracking}

\author{Fereshteh Aghaee Meibodi \\
	Department of Electrical and Computer Engineering\\
	University of Victoria \\
    3800 Finnerty Road \\
    Victoria, BC, Canada \\
	\texttt{fereshtehaghaee@uvic.ca} \\
	\And
	Amir Mehdi Soufi Enayati \\
    Department of Mechanical Engineering \\
    University of Victoria \\
    3800 Finnerty Road \\
    Victoria, BC, Canada \\
    \texttt{amsoufi@uvic.ca} \\
    \And
	Shadi Alijani \\
    Department of Mechanical Engineering \\
    University of Victoria \\
    3800 Finnerty Road \\
    Victoria, BC, Canada \\
    \texttt{shadialijani@uvic.ca} \\
    \And
	Homayoun Najjaran \\
    Department of Mechanical Engineering \\
    University of Victoria \\
    3800 Finnerty Road \\
    Victoria, BC, Canada \\
    \texttt{najjaran@uvic.ca} \\
}
\date{}

\renewcommand{\shorttitle}{\textit{arXiv} Template}

\hypersetup{
pdftitle={A template for the arxiv style},
pdfsubject={q-bio.NC, q-bio.QM},
pdfauthor={David S.~Hippocampus, Elias D.~Striatum},
pdfkeywords={First keyword, Second keyword, More},
}

\begin{document}
\maketitle

\begin{abstract}
Visual object tracking typically assumes that the initial template and subsequent search frames share the same sensing modality. In practice, sensor availability or operation may change over time, creating a substantial representation gap between template and search frames.
Unlike conventional multi-modal tracking where paired modalities are simultaneously available, cross-modal tracking requires localization when template and search frames originate from different active modalities.
Accordingly, we introduce \textbf{TSDA-Track}, a \textbf{T}emplate-\textbf{S}earch \textbf{D}omain \textbf{A}daptation framework to reduce modality discrepancy during training.
We investigate two feature alignment strategies. Pre-AFA TSDA-Track applies adversarial alignment before transformer's template-search interaction to suppress modality-specific bias. Enc-CFA TSDA-Track applies contrastive alignment to encoder representations after interaction to strengthen target-level cross-modal correspondence. Both variants retain a shared inference pipeline without modality-specific branches.
Experiments on LasHeR, and zero-shot evaluations on RGBT234 and GTOT under multiple cross-modal protocols demonstrate improvements over representative state-of-the-art trackers. For instance, under the modality-switch protocol on RGBT234, Pre-AFA TSDA-Track achieves an SR$_{0.5}$/PR of 43.2/56.0, compared with 36.8/50.0 for ToMP-101 baseline. In addition, a study on Anti-UAV-024 further verifies the applicability of TSDA-Track to aerial tracking. Our study highlights the effectiveness of feature alignment domain adaptation for cross-modal tracking.
\end{abstract}

\begin{figure}[!t]
  \centering
   \includegraphics[width=0.7\linewidth]{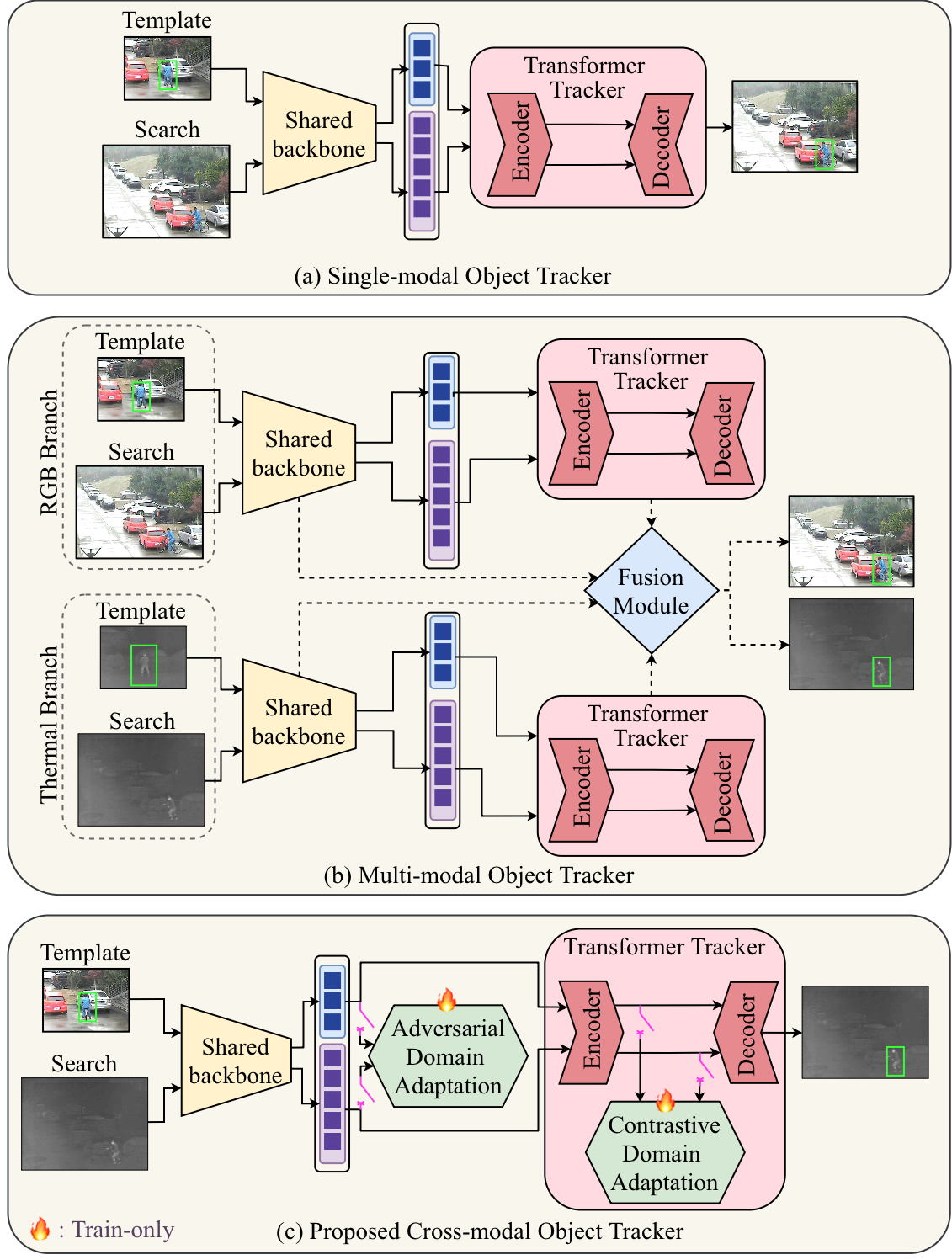}
   \caption{Comparison of (a) single-modal, (b) multi-modal RGB-T, and (c) proposed cross-modal tracking. Switches denote adaptive alignment strategy and interaction stage.}
   \label{fig:compare_tech}
\end{figure}

\section{Introduction}
Single object tracking (SOT) in computer vision aims to localize an arbitrary target throughout a video sequence given only its initial state in the first frame. Over the past decade, single-modal trackers, shown in Figure~\ref{fig:compare_tech}(a), have achieved remarkable progress on standard benchmarks, with various designs ranging from discriminative-based methods and Siamese networks to recent transformer-based architectures. However, they generally assume that the template and search frames share the same visual modality. To enhance robustness, multi-modal object trackers, such as RGB-T trackers, fuse complementary information from different sensors, shown in Figure~\ref{fig:compare_tech}(b). Although effective, they typically require all modalities to be simultaneously available and synchronized at each time step. Thus, both single-modal and conventional multi-modal tracking assume a consistent template-search modality configuration. This limits their applicability in modality switch scenarios, where the template and search frames may originate from different domains due to sensor failure, occlusion of one sensor, low illumination, or temporal switching. The resulting modality mismatch introduces a template-search domain gap that can weaken feature interaction and degrade target localization. This motivates the distinct setting of cross-modal object tracking, which aims to accurately localize a target when the template and search frames may originate from different sensing domains, such as RGB and thermal imagery. Previous efforts have addressed this scenario through modality-aware mechanisms, including modality prediction~\cite{liu2025prototype}, adaptive fusion weighting~\cite{liu2024cross, li2022cross}, and state-dependent routing~\cite{xu2025switrack}. These methods primarily select, weight, or route modality-dependent representations at inference time. Consequently, they do not explicitly optimize the template-search representation gap caused by sensing-domain mismatch, leaving feature interaction sensitive to the underlying distribution shift across domains.

To address this issue, this paper proposes two variants of TSDA-Track, a template-search domain adaptation framework that reduces cross-modal discrepancy through feature alignment, illustrated in Figure~\ref{fig:compare_tech}(c). Rather than relying on modality prediction, adaptive fusion, or state-dependent routing, TSDA-Track explicitly aligns template and search representations across modality domains. Specifically, we investigate two alignment strategies. Pre-AFA TSDA-Track applies adversarial domain adaptation~\cite{ganin2016domain} that encourages modality-invariant representations before template-search encoder interaction. Enc-CFA TSDA-Track applies contrastive feature alignment~\cite{oord2018representation,pang2021quasi} which promotes target-level cross-modal consistency after encoder. Feature alignment improves tracking robustness in dynamic environments with severe modality shift without requiring explicit modality branches at inference. A summary of our main contributions is:
\begin{itemize}
\item We formulate cross-modal tracking as a template-search domain mismatch perspective and evaluate it under fixed and temporally dynamic modality-switching protocols.
% RGB$\rightarrow$T, T$\rightarrow$RGB, and modality-switch protocols.
\item We propose TSDA-Track, a training-only domain-adaptation framework that reduces modality-induced template-search discrepancy through feature alignment. Since the alignment branches are removed after training, TSDA-Track retains a shared and modality-agnostic inference pipeline requiring no explicit modality estimation, adaptive fusion, modality-specific branches, or routing.
\item We develop two feature-alignment variants. Pre-AFA suppresses modality-specific bias via adversarial alignment, while Enc-CFA promotes target-level cross-modal consistency via contrastive alignment. Comprehensive ablations show that cross-modal performance depends on alignment objective, strength, and integration stage.
% , with adversarial alignment most effective before encoder interaction and contrastive alignment after encoder interaction.
\item TSDA-Track shows better overall performance compared with representative trackers across multiple cross-modal protocols on RGB-T datasets, including LasHeR and zero-shot RGBT234/GTOT evaluations, and is also validated on Anti-UAV aerial tracking.
\end{itemize}

\section{Related Work}
\paragraph{Single Object Tracking.}
SOT localizes an arbitrary target across video frames given its initial bounding-box. Existing SOT methods broadly fall into discriminative, Siamese-based, and transformer-based trackers~\cite{meibodi2025deep}. Discriminative trackers~\cite{danelljan2019atom, bhat2019learning} learn target-specific appearance models online, whereas Siamese-based trackers~\cite{yu2020deformable,liu2024siamdmu} formulate tracking as template-search matching using learned similarity functions. More recently, transformer-based trackers have advanced SOT through improved relation modeling. Early transformer-based trackers, such as TransT~\cite{chen2021transformer} and STARK~\cite{yan2021learning}, introduced attention mechanisms for robust feature fusion. ToMP~\cite{mayer2022transforming} reformulated discriminative model prediction with a transformer encoder-decoder and was later extended to multi-object tracking in TaMOs~\cite{mayer2024beyond}. OSTrack~\cite{ye2022joint} and MixFormer~\cite{cui2022mixformer} apply one-stream frameworks that jointly process template and search tokens. Recent advances include autoregressive sequence prediction in ARTrack~\cite{wei2023autoregressive} and SeqTrack~\cite{chen2023seqtrack}, spatio-temporal sequence modeling in ODTrack~\cite{zheng2024odtrack} and MCITrack~\cite{kang2025exploring}, and temporally-aware representation modeling in ROMTrack~\cite{cai2023robust}. However, most SOT trackers assume visually consistent, same-modality template and search frames, making them vulnerable to modality switches and template-search discrepancies.

\paragraph{Multi-Modal and Cross-Modal Object Tracking.}
To improve robustness under illumination variation, occlusion, and adverse imaging conditions, multi-modal tracking fuses complementary cues from different sensing modalities, addressing single-modal limitations. In RGB-T tracking, APFNet~\cite{xiao2022attribute} explores attribute-aware fusion and MTNet~\cite{hou2023mtnet} employs transformers to model global inter-modal relations and better exploit complementary cues. Recent RGB-X trackers, such as ProTrack~\cite{yang2022prompting}, ViPT~\cite{zhu2023visual}, and SDSTrack~\cite{hou2024sdstrack}, introduce prompt-based or unified adaptation mechanisms to improve multi-modal integration. However, these methods assume simultaneously available paired modalities and are not designed for modality-switching scenarios with only one active modality. Cross-modal object tracking relaxes this assumption by allowing modality-switching scenarios or template-search frames from different sensing domains. ProtoTrack~\cite{liu2025prototype} updates modality-specific prototypes through modality classification. MArMOT and MAFNet~\cite{li2022cross,liu2024cross} learn adaptive weights for RGB-NIR branches, and SwiTrack~\cite{xu2025switrack} applies tri-state routing. These cross-modal approaches rely on modality-aware inference mechanisms through modality prediction, adaptive weighting, or state-dependent routing. In contrast, the proposed Pre-AFA and Enc-CFA TSDA-Track directly reduce the template-search domain gap through training-only feature alignment, while retaining a shared inference pipeline without modality-specific branches or routing. 

\paragraph{Domain Adaptation for Tracking.}
Domain Adaptation reduces distribution shift between training and target domains. In visual tracking, it has mainly been explored to adapt models across environments or sensing domains. UDAT~\cite{ye2022unsupervised} aligns daytime and nighttime features via adversarial day-night discrimination, while subsequent work improves nighttime adaptation through transformer-based fusion~\cite{wei2024unsupervised}, temporal context alignment~\cite{fu2024prompt}, or SAM-assisted target-domain sample mining~\cite{fu2023sam}. For thermal tracking, PDAT~\cite{li2024progressive} transfers RGB knowledge to TIR by progressively aligning global and subdomain feature distributions, while EHDA~\cite{li2025efficient} performs hierarchical RGB-TIR adaptation at style and semantic levels. Trans-DAT~\cite{wu2024domain} further applies gradient reversal to reduce discrepancies among hyperspectral sensor domains. Unlike these methods that adapt trackers at the dataset or model level, the proposed TSDA-Track targets the within-sequence domain gap between cross-modal template and search frames.

\section{Proposed Method}

\begin{figure}[t!]
  \centering
   \includegraphics[width=1\linewidth]{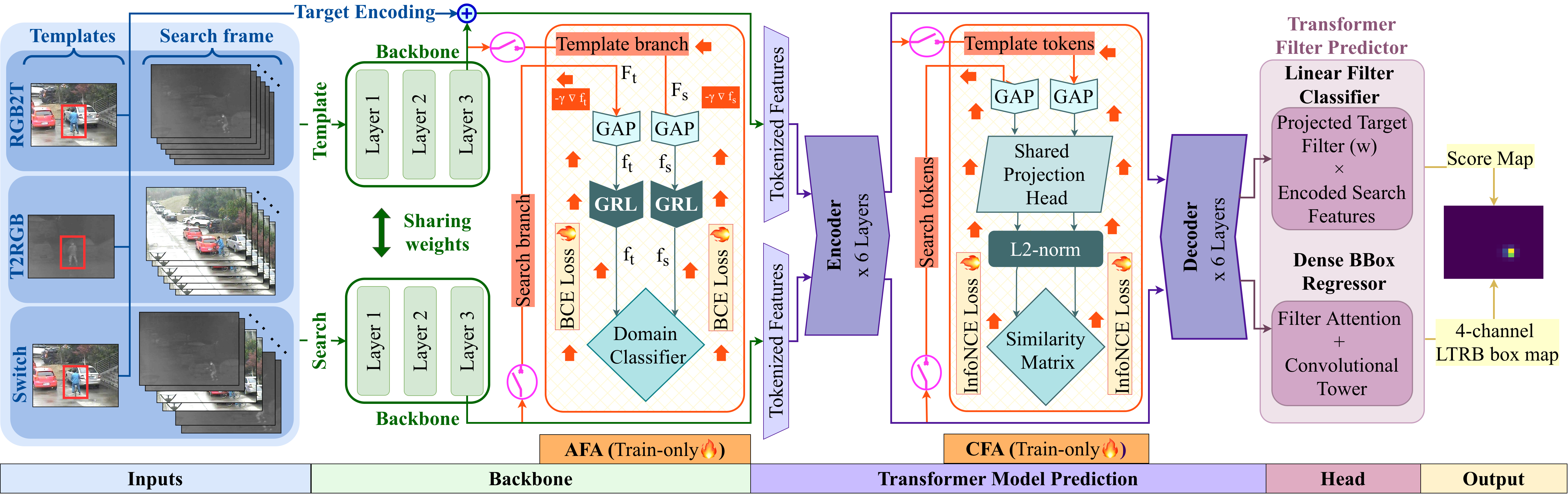}
   \caption{Overview of the proposed TSDA-Track for cross-modal tracking. AFA and CFA denote selectable offline-training alignment branches. Pre-AFA aligns pooled template and search features before the encoder via adversarial learning, whereas Enc-CFA aligns pooled encoder representations through contrastive learning.}
   \label{fig:tracking_fig}
\end{figure}

Figure~\ref{fig:tracking_fig} illustrates TSDA-Track, built on the transformer-based ToMP~\cite{mayer2022transforming} tracker. Given target-annotated templates $\mathcal{T}=\{{t_i}\}_{i=1}^{N_t}$ and a search frame $s$, a shared backbone extracts template and search features, respectively augmented with target-state and learnable test-frame encodings. A transformer encoder jointly processes these representations, and the decoder predicts target-specific model weights to estimate the target state in $s$. TSDA-Track addresses modality-mismatched tracking, where the template and search frames may originate from different sensing domains, including RGB and thermal imagery. Instead of modality-specific branches or routing, TSDA-Track applies training-only template-search feature-alignment to reduce modality-induced discrepancy while preserving the discriminative information for accurate tracking.

\subsection{Cross-Modal Tracking Setting}
\label{subsec:formulation}
Unlike standard SOT frameworks that assume same-modality template and search frames, cross-modal tracking allows them to originate from different sensing domains, such as RGB and thermal imagery, with distinct appearance statistics, contrast, and background cues. The resulting distribution shift can impair template-search interaction in the transformer and degrade target-state estimation. This motivates the feature-alignment objectives introduced below.

\subsection{Domain Adaptation for Cross-Modal Tracking}
To reduce template-search modality discrepancy while preserving target-discriminative features, TSDA-Track investigates two alternative training-only alignment strategies with distinct objectives and operating stages. Adversarial Feature Alignment (AFA) suppresses modality-specific cues before joint template-search encoding, promoting modality-invariant representations. Contrastive Feature Alignment (CFA) operates on jointly contextualized encoder outputs to strengthen target-conditioned cross-modal consistency. Each auxiliary alignment loss is jointly optimized with the tracking classification and regression losses and removed at inference. TSDA-Track thus retains the standard shared template-search pipeline without modality prediction,
modality-specific branches, or routing decisions.
\subsubsection{Adversarial Feature Alignment} \label{afa}
Adversarial domain adaptation is applied before the transformer encoder jointly processes template and search features. This provides a more compatible representation space for relation modeling because features at this stage may retain modality-specific appearance statistics that bias subsequent template-search interaction. Let $F^{\mathrm{pre}}_{t,i}$ and $F^{\mathrm{pre}}_{s}$ denote the pre-encoder feature maps of the $i$-th template and search frame, respectively. Since the template frames share the same modality during training, their pooled descriptors are aggregated as
\begin{equation}
f^{\mathrm{pre}}_{t}=
\frac{1}{N_t}\sum_{i=1}^{N_t}
\mathrm{GAP}\left(F^{\mathrm{pre}}_{t,i}\right), \ \ 
f^{\mathrm{pre}}_{s}=\mathrm{GAP}\left(F^{\mathrm{pre}}_{s}\right).
\end{equation}
where $\mathrm{GAP}(\cdot)$ denotes global average pooling. We apply adversarial alignment to these pooled descriptors rather than spatial feature maps because directly modifying spatial features may suppress the location-sensitive cues required for target localization in the tracking pipeline. Let $m_t,m_s\in{0,1}$ denote the modality labels of the template and search descriptors, where $0$ and $1$ correspond to RGB and thermal imagery, respectively. A shared pre-encoder domain discriminator $D_{\mathrm{pre}}$ predicts the modality of each descriptor. To adversarially train the shared feature extractor, a gradient-reversal layer (GRL)~\cite{ganin2015unsupervised} is inserted before the discriminator. The GRL acts as an identity mapping during the forward pass while multiplying the gradient propagated to the feature extractor by $-\gamma$ during backpropagation, resulting in $-\gamma\nabla{{f}^{\mathrm{pre}}_{x}} , x\in{t,s}$. Here, $\gamma$ controls the strength of the adversarial signal.
Thus, the discriminator distinguishes modalities while the shared feature extractor learns modality-invariant representations. The domain-classification loss is defined as
\begin{equation}
\mathcal{L}_{\mathrm{adv}} =
\frac{1}{2}
\sum_{x\in{t,s}}
\ell_{\mathrm{BCE}}
\left(
D_{\mathrm{pre}}(f^{\mathrm{pre}}_{x}),
m(x)
\right).
\end{equation}
where $\ell_{\mathrm{BCE}}$ denotes binary cross-entropy loss. The discriminator is optimized to identify the sensing modality, whereas the reversed gradient encourages the shared feature extractor to deceive it by making modality prediction difficult. This encourages modality-invariant representations and reduces modality-specific discrepancy before the transformer encoder jointly processes template and search information.

\subsubsection{Contrastive Feature Alignment} \label{cfa}
While Pre-AFA reduces modality discrepancy before template-search interaction, CFA regularizes jointly contextualized representations produced by the transformer encoder. At this stage, the encoder has already modeled interactions between target-conditioned template and current search representations. The resulting features therefore contain information directly relevant to subsequent target-model prediction. CFA exploits this information to encourage cross-modal semantic consistency between corresponding template-search pairs while preserving discrimination among different tracking tuples. Let $F^{\mathrm{enc}}_{t,i}$ and $F^{\mathrm{enc}}_{s}$ denote the post-encoder representations of the $i$-th template and search frame, respectively. Since template frames share the same modality during training, their global descriptors are aggregated as \begin{equation} f^{\mathrm{enc}}_{t} = \frac{1}{N_t} \sum_{i=1}^{N_t} \operatorname{GAP} \left( F^{\mathrm{enc}}_{t,i} \right), \ \ f^{\mathrm{enc}}_{s} = \operatorname{GAP} \left( F^{\mathrm{enc}}_{s} \right). \end{equation} We perform contrastive alignment on pooled encoder descriptors rather than enforcing token-wise spatial correspondence because RGB and thermal features can exhibit different local appearance and contrast responses. A shared projection head $P(\cdot)$ maps the pooled template and search descriptors into a shared embedding space, yielding projected template and search embeddings $z_t=P(f^{\mathrm{enc}}_{t})$ and $z_s=P(f^{\mathrm{enc}}_{s})$, respectively. For a mini-batch of $B$ tracking tuples, the $\ell_2$-normalized embeddings construct a batch-wise template-search similarity matrix:
\begin{equation}
\bar{z}_{t}^{\,i}
=
\frac{z_{t}^{\,i}}
{\left\|z_{t}^{\,i}\right\|_2},
\quad
\bar{z}_{s}^{\,j}
=
\frac{z_{s}^{\,j}}
{\left\|z_{s}^{\,j}\right\|_2},
\quad
i,j=1,\ldots,B.
\end{equation}
\begin{equation}
A_{ij}
=
\frac{
\operatorname{sim}
\left(
\bar{z}_{t}^{\,i},
\bar{z}_{s}^{\,j}
\right)
}{\tau}
=
\frac{
\bar{z}_{t}^{\,i\top}
\bar{z}_{s}^{\,j}
}{\tau},
\quad
A=[A_{ij}]_{i,j=1}^{B}.
\end{equation}
where $\operatorname{sim}(\cdot,\cdot)$ denotes cosine similarity and $\tau$ is a temperature parameter. Since the embeddings are normalized, their dot product equals cosine similarity. The diagonal entries $A_{ii}$ represent template and search embeddings from the same tracking tuple and are treated as positive pairs, whereas off-diagonal entries represent non-corresponding tuples in the mini-batch and serve as negatives. The template-to-search contrastive objective is defined as \begin{equation} \mathcal{L}_{t\rightarrow s} = -\frac{1}{B} \sum_{i=1}^{B} \log \frac{ \exp(A_{ii}) }{ \sum_{j=1}^{B} \exp(A_{ij}) }. \end{equation} Similarly, the search-to-template objective is computed by reversing the matching direction: \begin{equation} \mathcal{L}_{s\rightarrow t} = -\frac{1}{B} \sum_{i=1}^{B} \log \frac{ \exp(A_{ii}) }{ \sum_{j=1}^{B} \exp(A_{ji}) }. \end{equation} The final encoder-level contrastive loss is computed symmetrically as \begin{equation} \mathcal{L}_{\mathrm{ctr}}^{\mathrm{enc}} = \frac{1}{2} \left( \mathcal{L}_{t\rightarrow s} + \mathcal{L}_{s\rightarrow t} \right). \end{equation}
This bidirectional objective keeps each template representation close to its corresponding search-frame representation after encoder interaction while separating non-corresponding mini-batch samples. Jointly optimized with tracking losses, CFA promotes cross-modal consistency in encoder representations without discarding the discriminative information needed for target classification and bounding-box regression.

\subsection{Offline Training}
\label{}
Each training sample contains two target-annotated templates $\mathcal{T}=\{T_1,T_2\}$ and one search frame $s$. The templates are sampled from the same modality, $m(T_1)=m(T_2)$, to stabilize model prediction during training. The search frame is sampled from either the same or opposite modality with $p_{\mathrm{mismatch}}=0.5$, yielding balanced same-modality and cross-modal template-search pairs. For cross-modal samples, RGB-to-thermal and thermal-to-RGB template-search pairs are sampled uniformly.
% For cross-modal samples, RGB$\rightarrow$T and T$\rightarrow$RGB are selected uniformly.

\paragraph{Training objective.}
TSDA-Track augments the standard ToMP tracking objective with an auxiliary template-search alignment loss. The classification loss $\mathcal{L}_{\mathrm{cls}}$ supervises target score map using a Gaussian label~\cite{bhat2019learning}, while $\mathcal{L}_{\mathrm{GIoU}}$ supervises dense bounding-box predictions~\cite{mayer2022transforming}. The objectives for Pre-AFA and Enc-CFA are
\begin{equation}
\mathcal{L}_{\mathrm{AFA}} =
\lambda_{\mathrm{cls}}\mathcal{L}_{\mathrm{cls}} +
\lambda_{\mathrm{reg}}\mathcal{L}_{\mathrm{GIoU}} +
\lambda_{\mathrm{AFA}}\mathcal{L}_{\mathrm{adv}}^{\mathrm{pre}},
\end{equation}
\begin{equation}
\mathcal{L}_{\mathrm{CFA}} =
\lambda_{\mathrm{cls}}\mathcal{L}_{\mathrm{cls}} +
\lambda_{\mathrm{reg}}\mathcal{L}_{\mathrm{GIoU}} +
\lambda_{\mathrm{CFA}}\mathcal{L}_{\mathrm{ctr}}^{\mathrm{post}},
\end{equation}
where $\mathcal{L}_{\mathrm{adv}}^{\mathrm{pre}}$ is the pre-encoder modality-classification loss and $\mathcal{L}_{\mathrm{ctr}}^{\mathrm{post}}$ is the encoder-level contrastive loss. The weights $\lambda_{\mathrm{AFA}}$ and $\lambda_{\mathrm{CFA}}$ control the contribution of alignment losses.

\subsection{Online Tracking}
During online tracking, AFA and CFA branches are disabled, so TSDA-Track follows a shared and modality-agnostic ToMP inference pipeline without modality labels and modality-specific branches. Unlike approaches that explicitly identify or fuse modalities using inference modality classifiers or learned routing weights, TSDA-Track requires no such inference-time modules. This limits error propagation from inaccurate modality estimates, fusion weights, or routing and reduces sensitivity to whether the test modalities match those used to train inference-time modules. During online template memory update, the annotated initial template is retained while the remaining slots are updated with confident predictions. Thus, unlike offline training, the online memory may naturally contain RGB, thermal, or mixed-modality frames due to the input stream and update history.

\section{Experiments}

\subsection{Implementation Details}
We implement TSDA-Track by extending ToMP~\cite{mayer2022transforming}. The transformer contains six encoder and six decoder layers. All variants are implemented in Python with PyTorch,  Training is conducted on a single NVIDIA H100 GPU, and evaluation is performed on an NVIDIA TITAN RTX GPU.

\paragraph{Training and Inference Details.}
% We fine-tune the proposed models from a pretrained ToMP-101 checkpoint on the LasHeR~\cite{li2021lasher} training split for $30$ epochs using $40$k sampled subsequences and a batch size of $8$. Template and search crops are resized to $288\times288$, yielding $18\times18$ feature maps. The target filter size and search-area factor are set to $3$ and $5.0$, respectively. We use AdamW~\cite{loshchilov2017decoupled} with weight decay $10^{-4}$. Learning rates are $10^{-4}$ for the tracking head and alignment modules and $10^{-5}$ for ResNet layers, and reduced by a factor of $0.2$ after $20$ epochs. The backbone remains unfrozen to adapt the shared representation to the RGB-T distribution shift. Auxiliary alignment losses are enabled after five epochs to stabilize training. We set $\lambda_{\mathrm{cls}}=100$ and $\lambda_{\mathrm{reg}}=5$. For Pre-AFA TSDA-Track, $\lambda_{\mathrm{AFA}}=0.5$ and the GRL coefficient $\gamma$ is $0.1$. For Enc-CFA, $\lambda_{\mathrm{CFA}}=0.1$ and the $\tau$ is $0.1$. For the joint Pre-AFA+Enc-CFA variant, $\lambda_{\mathrm{AFA}}$ is $0.3$ with all other settings unchanged. Unlike ToMP, we independently apply target-centered center and scale jitter to both template and search crops to improve robustness under modality mismatch. All TSDA-Track variants use inference settings similar to ToMP. CM-ToMP-101 and CM-MAFNet denote ToMP-101 and MAFNet, respectively, fine-tuned with cross-modal template–search pairs but without the proposed explicit feature-alignment modules.

We fine-tune all models from a pretrained ToMP-101 checkpoint on the LasHeR~\cite{li2021lasher} training split for $30$ epochs using $40$k subsequences and a batch size of $8$. Template and search crops are resized to $288\times288$, yielding $18\times18$ feature maps. The target filter size and search-area factor are set to $3$ and $5.0$. We use AdamW~\cite{loshchilov2017decoupled} with weight decay $10^{-4}$, learning rates of $10^{-4}$ for the tracking head and alignment modules and $10^{-5}$ for ResNet layers, reduced by a factor of $0.2$ after $20$ epochs. The backbone is unfrozen to better align the representation mismatch through alignments. Auxiliary alignment losses are enabled after five epochs to stabilize training. We set $\lambda_{\mathrm{cls}}=100$ and $\lambda_{\mathrm{reg}}=5$. For Pre-AFA, $\lambda_{\mathrm{AFA}}=0.5$ and the GRL coefficient $\gamma$ is $0.1$. For Enc-CFA, $\lambda_{\mathrm{CFA}}=0.1$ and $\tau$ is $0.1$. In the joint variant, $\lambda_{\mathrm{AFA}}$ is $0.3$ and $\lambda_{\mathrm{CFA}}$ is $0.1$ with all other settings unchanged. Unlike ToMP, independent target-centered position and scale jitter are applied to both template and search crops to improve robustness under modality mismatch. All TSDA-Track variants use ToMP inference settings. CM-ToMP-101 and CM-MAFNet denote ToMP-101 and MAFNet, respectively, fine-tuned with cross-modal template–search pairs but without the proposed explicit feature-alignment modules. 
\paragraph{Experimental Setup}
% We conduct ablation studies on RGBT234~\cite{li2019rgb} and compare TSDA-Track with representative single-object trackers on RGBT234, the LasHeR~\cite{li2021lasher} test split, and GTOT~\cite{li2016learning}. To further assess applicability to aerial RGB-T tracking, we train separate models on the Anti-UAV-024~\cite{jiang2021anti} training split and evaluate them on its test split. 
We design three novel template-search modality-discrepancy protocols to assess robustness to fixed and dynamic modality shifts. RGB$\rightarrow$T initializes the tracker with an RGB template and thermal search frames, whereas T$\rightarrow$RGB reverses this setting. In Switch mode, the template is RGB and the search modality alternates between thermal and RGB every $200$ frames. We report success rate at IoU $0.5$ (SR$_{0.5}$) and precision rate at a $20$-pixel center-error threshold (PR) (see more SR/PR plots in suppl. material).
% Formal metric definitions are provided in the supplementary material.

\subsection{Ablation Study and Analysis}
% We analyze the main TSDA-Track design choices and their effect on cross-modal robustness. The CM-ToMP-101 denotes ToMP-101 which is trained with cross-modal template-search pairs but without explicit feature alignment. The TSDA-Track variants are then analyzed by varying the alignment objective, its application stage, and alignment strength.
We conduct an ablation study of TSDA-Track to examine how the alignment objective, its stage-specific integration, and alignment strength affect cross-modal robustness.
% Compared with ToMP-101, this yields gains of $8.8/9.0$ points for RGB$\rightarrow$T and $6.4/6.0$ points for Switch. 
% RGBT234~\cite{li2019rgb}

\begin{table}[b]
\centering
\caption{Zero-shot ablation on cross-modal training (CM), alignment strategy and stage on RGBT234~\cite{li2019rgb}. Adv./Con.\ FA denote adversarial/contrastive alignment. Pre and Post refer to pre- and post-encoder alignment. The top two results are bolded and underlined, respectively.}
\label{tab:ablation_strategy_stage}
\footnotesize
\renewcommand{\arraystretch}{1.05}
\setlength{\tabcolsep}{2.2pt}
\begin{adjustbox}{max width=\columnwidth}
\begin{tabular}{@{}lccc*{6}{c}@{}}
\toprule
\multirow{2}{*}{Method}
& \multirow{2}{*}{\makecell{CM\\Train.}}
& \multirow{2}{*}{\makecell{Adv.\\FA}}
& \multirow{2}{*}{\makecell{Con.\\FA}}
& \multicolumn{2}{c}{RGB$\rightarrow$T}
& \multicolumn{2}{c}{T$\rightarrow$RGB}
& \multicolumn{2}{c}{Switch}
\\
\cmidrule(lr){5-6}
\cmidrule(lr){7-8}
\cmidrule(lr){9-10}
& & &
& SR$_{0.5}$ & PR
& SR$_{0.5}$ & PR
& SR$_{0.5}$ & PR
\\
\midrule

ToMP-101
& $-$&$-$ &$-$
& 31.7 & 44.9
& 33.2 & 43.1
& 36.8 & 50.0
\\
CM-ToMP-101
& \cmark & $-$&$-$
& 34.8 & 47.8
& 36.4 & 47.3
& 35.9 & 47.6
\\[-1pt]
\addlinespace[1pt]
\rowcolor{black!10}
\multicolumn{10}{@{}l@{}}{\strut\textbf{TSDA-Track variants:}}
\\[-1pt]
\addlinespace[1pt]
Pre-AFA
& \cmark
% & \makecell{\cmark\\[-2pt]\scriptsize Pre}
& Pre
& $-$
& \textbf{40.5} & \textbf{53.9}
& \underline{39.9} & \underline{52.4}
& \textbf{43.2} & \textbf{56.0}
\\
Pre-CFA
& \cmark
& $-$
% & \makecell{\cmark\\[-2pt]\scriptsize Pre}
& Pre
& 37.0 & 48.1
& 35.9 & 46.1
& 38.6 & 49.4
\\
Enc-AFA
& \cmark
% & \makecell{\cmark\\[-2pt]\scriptsize Post}
& Post
&$-$
& 36.6 & 50.6
& 37.5 & 50.3
& 38.2 & 51.7
\\
Enc-CFA
& \cmark
&$-$
% & \makecell{\cmark\\[-2pt]\scriptsize Post}
& Post
& 38.6 & \underline{52.0}
& 38.6 & 51.9
& 41.5 & \underline{54.4}
\\
Pre-AFA + Enc-CFA
& \cmark
% & \makecell{\cmark\\[-2pt]\scriptsize Pre}
& Pre
% & \makecell{\cmark\\[-2pt]\scriptsize Post}
& Post
& \underline{39.7} & 50.7
& \textbf{42.6} & \textbf{53.2}
& \underline{42.6} & 53.6
\\
\bottomrule
\end{tabular}
\end{adjustbox}
\end{table}
\subsubsection{Effect of Alignment Strategy and Stage}
Table~\ref{tab:ablation_strategy_stage} evaluates the effects of cross-modal training, alignment strategy, and alignment stage. Compared with ToMP-101, CM-ToMP-101 improves the fixed RGB$\rightarrow$T and T$\rightarrow$RGB settings but degrades Switch performance, showing that cross-modal training alone is insufficient for repeated modality changes. Explicit feature alignment improves robustness across protocols. Pre-AFA achieves the best RGB$\rightarrow$T and Switch results, $40.5/53.9$ and $43.2/56.0$, respectively. Its advantage over Enc-AFA indicates that adversarial alignment is most effective before transformer interaction, where modality-specific appearance bias can be suppressed before template-search relation modeling in the encoder. In contrast, CFA is more effective after encoder interaction. Early features remain strongly influenced by modality-specific appearance and local contrast, making Pre-CFA less reliable for target-level alignment. Enc-CFA instead operates on contextualized template-search representations after encoder, where positive pairs better represent target-level correspondence and negatives provide more meaningful discriminative supervision. Pre-AFA remains stronger because it prevents modality bias from entering transformer interaction, whereas Enc-CFA cannot fully recover relation modeling already affected by modality mismatch. Combining the strongest individual configurations, Pre-AFA+Enc-CFA achieves the best T$\rightarrow$RGB result ($42.6/53.2$) but does not consistently improve RGB$\rightarrow$T or Switch. Thus, the objectives are not universally additive. Joint optimization may over-constrain the shared representation and attenuate target-discriminative cues required for accurate localization during tracking.

\subsubsection{Effect of Alignment Strength}
Table~\ref{tab:ablation_strength} evaluates alignment strength on TSDA-Track variants. In Pre-AFA, the GRL coefficient scales the adversarial gradient on the shared feature extractor. $\gamma=0.1$ consistently outperforms $\gamma=1.0$, indicating that excessive adversarial alignment can suppress target-discriminative features needed for tracking. In Enc-CFA, a lower temperature $\tau$ sharpens the similarity distribution and increases contrastive pressure. $\tau=0.1$ performs best across protocols, suggesting that moderate contrastive alignment better accommodates modality and appearance variation.

\begin{table}[!tb] 
\centering 
\caption{Zero-shot evaluation of the effect of Pre-AFA and Enc-CFA alignment strengths on RGBT234~\cite{li2019rgb}.} \label{tab:ablation_strength} \renewcommand{\arraystretch}{1.05} \setlength{\tabcolsep}{2.2pt} \begin{adjustbox}
{max width=\columnwidth} 
\begin{tabular}{@{}lcccccccc@{}} 
\toprule \multirow{2}{*}{Method} & 
\multirow{2}{*}{Parameter} & 
\multirow{2}{*}{Strength} & \multicolumn{2}{c}{RGB$\rightarrow$T} & \multicolumn{2}{c}{T$\rightarrow$RGB} & \multicolumn{2}{c}{Switch}\\ \cmidrule(l{2pt}r{2pt}){4-5} \cmidrule(l{2pt}r{2pt}){6-7} \cmidrule(l{2pt}r{2pt}){8-9} & & &  SR$_{0.5}$ & PR & SR$_{0.5}$ & PR & SR$_{0.5}$ & PR\\
\midrule
{\multirow{2}{*}{Pre-AFA TSDA-Track}} &
\multirow{2}{*}{$\gamma$} &
$0.1$ & \textbf{40.5} & \textbf{53.9} & \textbf{39.9} & \textbf{52.4} & \textbf{43.2} & \textbf{56.0} \\ \cmidrule(l{6pt}r{2pt}){3-9}
 & & 1 & 38.5 & 49.5 & 39.7 & 49.7 & 40.7 & 51.5 \\ 
\midrule
{\multirow{2}{*}{Enc-CFA TSDA-Track}} &
\multirow{2}{*}{$\tau$} &
0.05 & 36.8 & 49.4 & 37.2 & 47.4 & 39.1 & 50.7  \\
\cmidrule(l{6pt}r{2pt}){3-9}
 & & 0.1 &\textbf{ 38.6} & \textbf{52.0}
& \textbf{38.6} & \textbf{51.9}
& \textbf{41.5} & \textbf{54.4}  \\
\bottomrule 
\end{tabular} 
\end{adjustbox} 
\end{table} 

\subsubsection{Feature-Level Analysis}
\begin{figure}[!tb]
  \centering
  \includegraphics[width=0.7\linewidth]{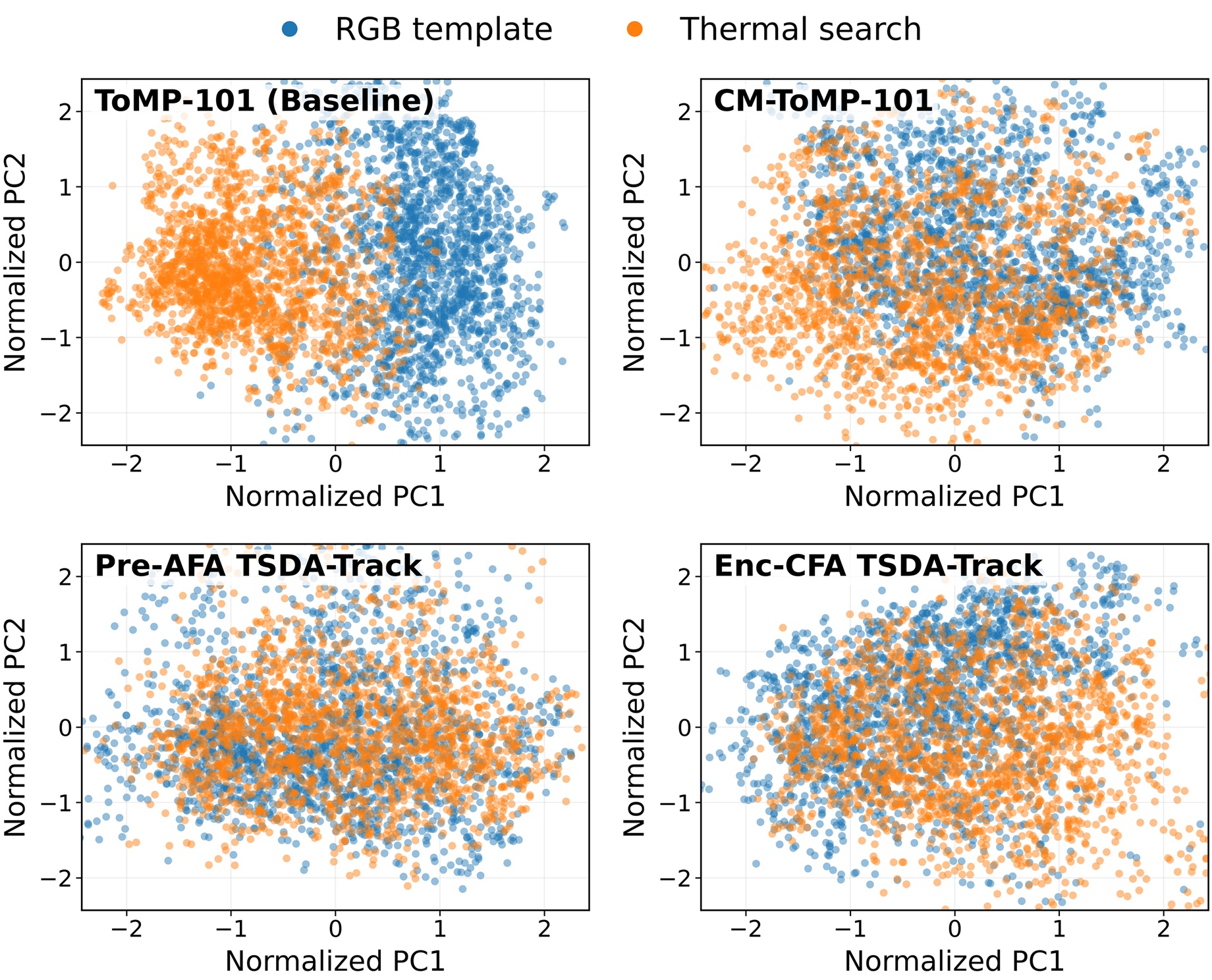}
  \caption{PCA visualization of normalized pre-encoder representations on RGBT234~\cite{li2019rgb}. 
  % Greater overlap indicates reduced modality-specific separation.
  }
  \label{fig:pca_pre}
\end{figure}

Figure~\ref{fig:pca_pre} visualizes normalized pre-encoder representations of RGB templates and thermal searches on RGBT234. ToMP-101 exhibits clear modality separation, indicating a substantial pre-interaction RGB-thermal gap. CM-ToMP-101 increases overlap through cross-modal training but retains modality-specific structure. Pre-AFA TSDA-Track yields the strongest mixing by directly suppressing pre-encoder modality cues before transformer interaction. Enc-CFA also reduces the gap, but the distributions remain more structured as it promotes post-encoder target correspondence rather than pre-encoder modality invariance. 

To complement the PCA visualization, Table~\ref{tab:feature_ratio_analysis} quantifies identity-preserving cross-modal alignment at the pre- and post-encoder levels. For each RGB-template/thermal-search pair $i$, $d_{i,l}^{+}$ denotes the cosine distance between matched features from the same sequence, and $d_{i,l}^{-}$ denotes the mean distance to unmatched thermal-search features from other sequences in the batch. We define
\begin{equation}
R_{loc} =
\operatorname{median}_{i}
\left(
\frac{d_{i,l}^{+}}{d_{i,l}^{-}+\epsilon}
\right),
\quad loc \in \{\mathrm{pre},\mathrm{post}\},
\label{eq:pos_neg_ratio}
\end{equation}
where $\epsilon$ is a small constant. Lower $R_{loc}$ indicates stronger same-object RGB-thermal alignment relative to different-object pairs. CM-ToMP-101 reduces both ratios compared with ToMP-101, confirming the benefit of cross-modal training. Pre-AFA achieves the lowest pre-encoder ratio ($0.435$), consistent with its direct pre-interaction alignment objective, whereas Enc-CFA achieves the lowest post-encoder ratio ($0.594$), matching its encoder-level contrastive supervision. These results quantitatively support that each alignment strategy is most effective at its intended feature stage.

\begin{table}[!b]
\centering
\caption{Positive/negative cross-modal distance ratio on RGBT234~\cite{li2019rgb}. Lower values indicate stronger same-object alignment relative to different-object pairs.}
\label{tab:feature_ratio_analysis}
\renewcommand{\arraystretch}{1.15}
\setlength{\tabcolsep}{2.2pt}
\begin{adjustbox}{max width=\columnwidth}
\begin{tabular}{@{}lcc@{}}
\toprule
\multirow{1}{*}{Method} & \makecell{Pre-encoder \\ $R_{\mathrm{pre}}\downarrow$} & \makecell{Post-encoder \\ $R_{\mathrm{post}}\downarrow$ }\\
\midrule
Baseline:ToMP-101~\cite{mayer2022transforming} & 0.730 & 0.862 \\
CM-ToMP-101 & 0.496 & 0.625 \\
Pre-AFA TSDA-Track & \textbf{0.435} & 0.625 \\
Enc-CFA TSDA-Track & 0.474 & \textbf{0.594} \\
\bottomrule
\end{tabular}
\end{adjustbox}
\end{table}

\subsection{Comparison with Representative Trackers}
\begin{table*}[!tb]
\centering
\caption{Cross-modal evaluation against representative single-object trackers on RGBT234~\cite{li2019rgb}, LasHeR~\cite{li2021lasher}, and GTOT~\cite{li2016learning}  under RGB$\rightarrow$T, T$\rightarrow$RGB, and Switch protocols. RGBT234 and GTOT results are zero-shot. Bold, underlined, and italic values indicate the first-, second-, and third-best results, respectively.
}
\label{tab:sota_crossmodal}
\renewcommand{\arraystretch}{1.05}
\setlength{\tabcolsep}{2.2pt}
\begin{adjustbox}{max width=\textwidth}
\begin{tabular}{l*{18}{c}}
\toprule
\multirow{3}{*}{Methods}
& \multicolumn{6}{c}{RGBT234~\cite{li2019rgb}}
& \multicolumn{6}{c}{LasHeR\cite{li2021lasher}}
& \multicolumn{6}{c}{GTOT~\cite{li2016learning}} 
\\
\cmidrule(l{2pt}r{4pt}){2-7}
\cmidrule(l{6pt}r{4pt}){8-13}
\cmidrule(l{6pt}r{2pt}){14-19}
& \multicolumn{2}{c}{RGB$\rightarrow$T}
& \multicolumn{2}{c}{T$\rightarrow$RGB}
& \multicolumn{2}{c}{Switch}{\hskip 5pt} 
& \multicolumn{2}{c}{RGB$\rightarrow$T}
& \multicolumn{2}{c}{T$\rightarrow$RGB}
& \multicolumn{2}{c}{Switch}
&  \multicolumn{2}{c} {RGB$\rightarrow$T}
& \multicolumn{2}{c}{T$\rightarrow$RGB}
& \multicolumn{2}{c}{Switch} \\
\cmidrule(l{2pt}r{2pt}){2-3} \cmidrule(l{2pt}r{2pt}){4-5} \cmidrule(l{2pt}r{4pt}){6-7}
\cmidrule(l{6pt}r{2pt}){8-9} \cmidrule(l{2pt}r{2pt}){10-11} \cmidrule(l{2pt}r{4pt}){12-13}
\cmidrule(l{6pt}r{2pt}){14-15} \cmidrule(l{2pt}r{2pt}){16-17} \cmidrule(l{2pt}r{2pt}){18-19}
& SR$_{0.5}$ & PR
& SR$_{0.5}$ & PR
& SR$_{0.5}$ & PR 
& {\hskip 4pt} SR$_{0.5}$ & PR
& SR$_{0.5}$ & PR
& SR$_{0.5}$ & PR 
& {\hskip 4pt} SR$_{0.5}$ & PR
& SR$_{0.5}$ & PR
& SR$_{0.5}$ & PR \\
\midrule
ATOM~\cite{danelljan2019atom}
& 27.7 & 37.6 & 29.5 & 39.7 & 32.0 & 41.2
& 14.7 & 18.7 & 14.2 & 19.5 & 17.7 & 20.7
& 23.9 & 39.4 & 38.8 & 52.3 & 27.9 & 39.6 \\
DiMP-50~\cite{bhat2019learning}
& 24.9 & 32.8 & 32.9 & 44.0 & 33.3 & 41.3
& 15.3 & 17.3 & 18.4 & 22.6 & 19.6 & 22.4
& 27.3 & 38.7 & 30.0 & 45.0 & 27.4 & 40.0 \\
KeepTrack~\cite{mayer2021learning}
& 26.0 & 34.2 & 29.2 & 36.8 & 36.7 & 44.8
& 16.6 & 18.8 & 18.0 & 20.4 & 21.6 & 23.9
& 28.1 & 41.2 & 25.7 & 37.0 & 29.1 & 41.6 \\
STARK-ST~\cite{yan2021learning}
& 19.7 & 24.6 & 21.6 & 26.7 & 30.6 & 35.8
& 11.3 & 12.8 & 16.7 & 18.0 & 18.2 & 20.0
& 31.6 & 44.7 & 26.0 & 33.4 & 34.1 & 46.8 \\
ToMP-101~\cite{mayer2022transforming}
& 31.7 & 44.9 & 33.2 & 43.1 & 36.8 & 50.0
& 21.6 & 25.7 & 19.4 & 24.3 & 25.2 & 29.3
& 40.7 & 60.4 & 37.7 & 55.6 & 39.9 & 59.9 \\
OSTrack~\cite{ye2022joint}
& 19.1 & 25.3 & 29.7 & 36.6 & 30.1 & 36.0
& 10.2 & 11.8 & 21.2 & 23.2 & 18.0 & 19.8
& 30.0 & 37.8 & 32.0 & 43.9 & 32.9 & 40.8 \\
MixFormer-l~\cite{cui2022mixformer}
& 25.6 & 31.9 & 29.4 & 36.3 & 33.7 & 40.1
& 18.4 & 20.3 & 23.4 & 26.9 & 24.4 & 26.5
& 37.2 & 55.2 & 38.6 & 47.9 & 38.9 & 56.8 \\
ODTrack-b~\cite{zheng2024odtrack}
& 25.4 & 42.6 & \textcolor{black}{\textbf{44.0}} & \textcolor{black}{\underline{55.7}} & 33.4 & 47.5
& 14.9 & 19.7 & \underline{\textcolor{black}{30.4}} & \textit{\textcolor{black}{35.5}} & 19.4 & 23.6
& 37.8 & 54.4 & \textit{40.9} & 52.2 & 39.6 & 56.4 \\
SeqTrack-l384~\cite{chen2023seqtrack}
& 27.4 & 37.0 & 32.1 & 40.4 & \textit{\textcolor{black}{36.8}} & 45.1  
& 19.4 & 22.4 & 24.1 & 28.8 & 26.3 & 29.7
& 33.3 & 54.7 & 32.1 & 47.6 & 34.3 & 55.2 \\
MCITrack-l384~\cite{kang2025exploring}
& 22.6 & 27.8 & 25.9 & 31.1 & 34.3 & 39.3
& 12.7 & 13.4 & 23.4 & 25.3 & 21.0 & 21.8
& 32.4 & 42.6 & 24.5 & 32.1 & 35.4 & 45.8 \\
MAFNet~\cite{liu2024cross}
& 20.8 & 36.8 & 25.2 & 45.7 & 24.2 & 41.3
& 13.1 & 20.3 & 14.6 & 22.5 & 14.3 & 21.7
& 25.3 & 47.4 & 26.3 & 60.8 & 25.2 & 47.7 \\
% \midrule
\rowcolor{black!10}
\multicolumn{19}{@{}l@{}}{\strut\textbf{Cross-modal trained:}}
\\[-1pt]
\addlinespace[1pt]

CM-MAFNet
& \textcolor{black}{32.0}&  \textcolor{black}{\textbf{54.7}} &  \textcolor{black}{30.4} &  \textcolor{black}{\textbf{57.2}} &  \textcolor{black}{34.9} &  \textcolor{black}{\textbf{58.3}}
&  \textcolor{black}{21.8} &  \textcolor{black}{\underline{33.6}} &  \textcolor{black}{23.4} &  \textcolor{black}{\underline{36.8}} &  \textcolor{black}{22.5} &  \textcolor{black}{\underline{35.6}}
&  \textcolor{black}{28.7} &  \textcolor{black}{52.7} &  \textcolor{black}{37.3} &  \textcolor{black}{\textbf{72.0}} &  \textcolor{black}{31.0} &  \textcolor{black}{55.3} \\
CM-ToMP-101         
& \textit{\textcolor{black}{34.8}} & \textcolor{black}{47.8}
& 36.4 & 47.3
& 35.9 & \textcolor{black}{47.6} & \textit{\textcolor{black}{26.3}} &\textcolor{black}{30.3} & \textit{\textcolor{black}{29.6}} & \textcolor{black}{32.4} & \textit{\textcolor{black}{27.3}} & \textcolor{black}{31.2}
& \underline{\textcolor{black}{47.3}} & \underline{\textcolor{black}{62.8}} & \textcolor{black}{39.3} & \textcolor{black}{56.4} & \underline{\textcolor{black}{46.3}} & \underline{\textcolor{black}{62.0}} \\
Enc-CFA TSDA-Track            
& \underline{\textcolor{black}{38.6}} & \textit{\textcolor{black}{52.0}}
& \textit{\textcolor{black}{38.6}} & \textcolor{black}{51.9}
& \underline{\textcolor{black}{41.5}} & \textit{\textcolor{black}{54.4}}
& \underline{\textcolor{black}{28.4}} & \textit{\textcolor{black}{32.8}} & 28.8 & \textcolor{black}{32.4} & \underline{\textcolor{black}{28.3}} & \textit{\textcolor{black}{32.9}}
& \textbf{\textcolor{black}{49.4}} & \textbf{\textcolor{black}{65.3}} & \textbf{\textcolor{black}{53.7}} & \textit{\textcolor{black}{67.6}} & \textbf{\textcolor{black}{49.6}} & \textbf{\textcolor{black}{65.2}}  \\
Pre-AFA TSDA-Track
& \textbf{\textcolor{black}{40.5}} & \underline{\textcolor{black}{53.9}}
& \underline{\textcolor{black}{39.9}} & \textit{\textcolor{black}{52.4}}
& \textbf{\textcolor{black}{43.2}}& \underline{\textcolor{black}{56.0}}
& \textbf{\textcolor{black}{36.6}} & \textbf{\textcolor{black}{38.9}} & \textbf{\textcolor{black}{37.7}} & \textbf{\textcolor{black}{40.0}} & \textbf{\textcolor{black}{37.3}} & \textbf{\textcolor{black}{39.8}}
& \textit{\textcolor{black}{45.7}} & \textit{\textcolor{black}{61.3}} &\underline{\textcolor{black}{52.9}} &\underline{\textcolor{black}{70.0}} & \textit{\textcolor{black}{45.8}} & \textit{\textcolor{black}{61.8}} \\
\bottomrule
\end{tabular}
\end{adjustbox}
\end{table*}

% Table~\ref{tab:sota_crossmodal} compares TSDA-Track with representative SOT frameworks and cross-modal MAFNet~\cite{liu2024cross} tracker using the provided models from their authors under the proposed RGB-T template-search modality-discrepancy protocols. Direct comparison with other cross-modal trackers is currently not feasible due to unavailable public implementations. 
% % These methods also target RGB-NIR data under real hardware-triggered modality switching, a different sensor pair and mechanism from the RGB-T template-search mismatch protocol studied here. 
% The results show that strong single-modality tracking does not necessarily transfer to cross-modal tracking, and that cross-modal training alone is insufficient for stable performance under modality discrepancy. RGBT234 and GTOT are evaluated zero-shot, without target-dataset training or fine-tuning.
% Direct comparison with other cross-modal trackers is currently infeasible because their public implementations are unavailable.
Table~\ref{tab:sota_crossmodal} compares TSDA-Track with representative SOT trackers and the available cross-modal MAFNet~\cite{liu2024cross} using their released checkpoints under the proposed RGB-T template-search modality-discrepancy protocols.
Note that comparison with other cross-modal trackers is infeasible because of unavailable implementations or models.
% MAFNet’s RGB-NIR-trained modality-specific branches with adaptive fusion weights transfer poorly to thermal imagery and underperform several single-modal trackers, showing that cross-modal fusion learned for one sensor pair may not generalize to another without adaptation. The substantial gains of CM-MAFNet over MAFNet confirm the importance of sensor-pair adaptation. More broadly, the results show that strong single-modal tracking does not necessarily transfer to cross-modal settings. CM-ToMP-101 and CM-MAFNet results further indicate that cross-modal training alone is insufficient for stable performance under modality discrepancy. RGBT234 and GTOT are evaluated zero-shot, without target-dataset training or fine-tuning.
% RGB$\rightarrow$T is challenging because thermal search frames lack the color and fine-grained texture cues of the RGB template, whereas T$\rightarrow$RGB provides richer search details but still requires cross-modal matching. Switch further evaluates robustness to temporal modality changes and confidence-gated mixed-modality memory updates.
MAFNet’s RGB-NIR-trained modality-specific branches and its adaptive fusion weights transfer poorly to RGB-thermal, causing it to underperform several single-modal trackers. Therefore, cross-modal fusion weights learned for one sensor pair may not generalize to another without adaptation, as the improvement of CM-MAFNet over MAFNet further confirms the importance of sensor-pair adaptation. In addition, strong single-modal tracking does not necessarily transfer to cross-modal settings, and the CM-ToMP-101 and CM-MAFNet results further indicate that cross-modal training alone is insufficient for stable performance under modality discrepancy. RGBT234 and GTOT are evaluated zero-shot, without target-dataset training or fine-tuning.
% \paragraph{RGBT234~\cite{li2019rgb}.} Pre-AFA TSDA-Track achieves the best zero-shot performance in RGB$\rightarrow$T and Switch with $40.5/53.9$ and $43.2/56.0$ in SR$_{0.5}$/PR, respectively. These results indicate that suppressing modality bias before transformer interaction is beneficial where modality discrepancy is coupled with diverse target appearance and background changes. ODTrack achieves the best T$\rightarrow$RGB result ($44.0/55.7$) benefiting from richer RGB search stream, but its lower RGB$\rightarrow$T and Switch scores show that this advantage is direction-specific rather than consistent cross-modal robustness. In contrast, Pre-AFA maintains strong performance under repeated modality switches by stabilizing target representations as the mixed-modality memory evolves. Qualitative comparisons in Figure~\ref{fig:qualitative_comparison} support the quantitative findings, showing that explicit template-search feature alignment improves localization across modality transitions.

\begin{figure}[!tb]
    \centering
    %==================== (a) Success row ====================
    \begin{subfigure}[t]{\columnwidth}
        \centering
        \includegraphics[width=1\linewidth]{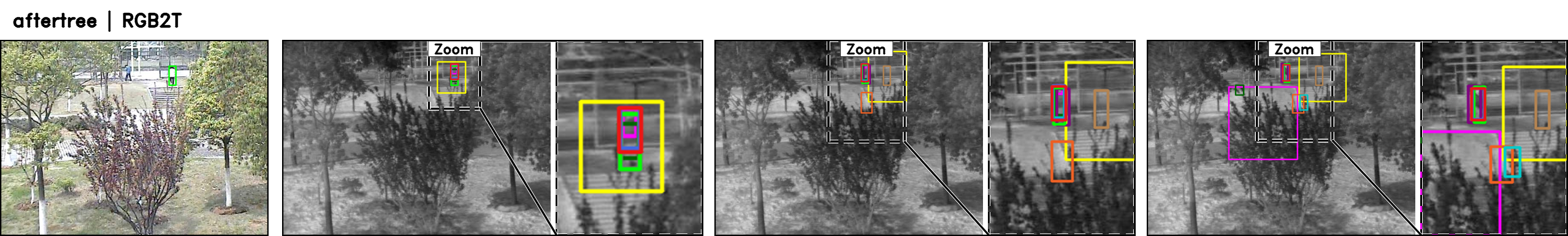}
    \end{subfigure}
    \begin{subfigure}[t]{\columnwidth}
        \centering
        \includegraphics[width=1.0\linewidth]{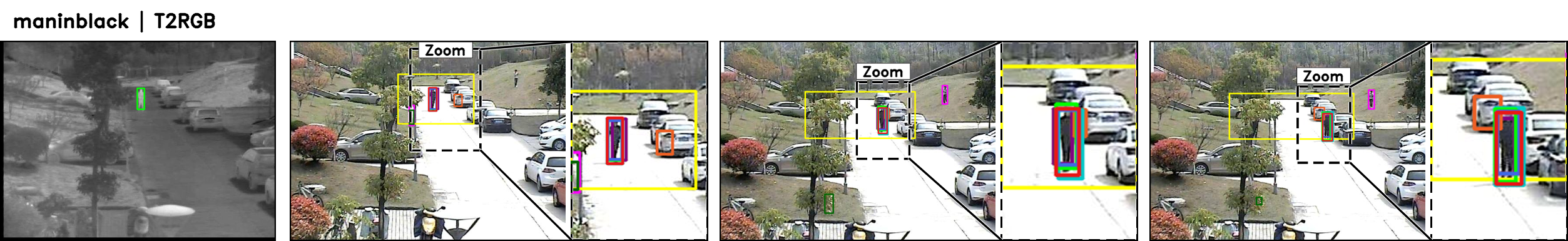}
    \end{subfigure}
    \begin{subfigure}[t]{\columnwidth}
        \centering
        \includegraphics[width=1.0\linewidth]{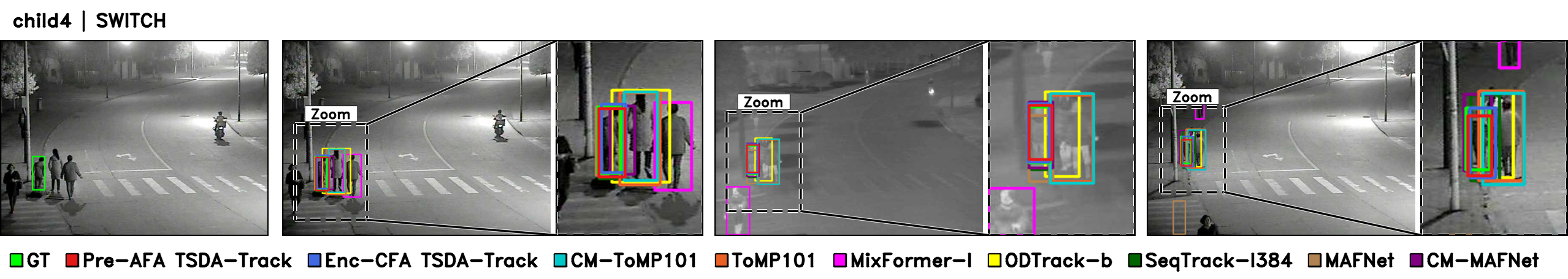}
    \end{subfigure}
    \caption{Qualitative comparison on RGBT234~\cite{li2019rgb} under RGB→T, T→RGB, and Switch protocols.}
    \label{fig:qualitative_comparison}
\end{figure}

\paragraph{RGBT234~\cite{li2019rgb}.} Pre-AFA TSDA-Track achieves the highest zero-shot SR performance in RGB$\rightarrow$T and Switch with $40.5$ and $43.2$, respectively. This indicates that suppressing modality bias before transformer interaction is beneficial where modality discrepancy is coupled with diverse target appearance and background changes. Compared with MAFNet, CM-MAFNet improves after RGB-T cross-modal fine-tuning and achieves high PR across protocols. However, its substantial lower SR indicates less consistent bounding-box overlap. ODTrack achieves the highest T$\rightarrow$RGB SR of $44.0$ benefiting from richer RGB search stream, but its lower RGB$\rightarrow$T and Switch scores show that this advantage is direction-specific rather than consistent cross-modal robustness. Qualitative comparisons in Figure~\ref{fig:qualitative_comparison} support the quantitative findings that explicit template-search feature alignment improves tracking across modality transitions.

\paragraph{LasHeR~\cite{li2021lasher}.}  Pre-AFA TSDA-Track achieves the best performance across all protocols on LasHeR. Its consistent advantage under modality discrepancies indicates that pre-encoder alignment effectively suppresses modality bias before template-search interaction, yielding more robust target modeling where cross-modal discrepancy is coupled with LasHeR’s substantial appearance variation.

\paragraph{GTOT~\cite{li2016learning}.} In zero-shot evaluation on GTOT, Enc-CFA TSDA-Track achieves the best SR$_{0.5}$ under all three protocols. CM-MAFNet obtains the highest T$\rightarrow$RGB PR of $72.0$ but a substantially lower SR of $37.3$, compared with $53.7$ for Enc-CFA. GTOT trend differs from RGBT234 and LasHeR, where Pre-AFA is stronger. GTOT differs from the training data in acquisition conditions and exhibits a more diverse modality distribution, as it also includes grayscale input. This shift may limit Pre-AFA transferability, as its pre-encoder alignment can remain sensitive to input-domain statistics. In contrast, Enc-CFA aligns target-conditioned template-search representations after encoder interaction, with less explicit reliance on modality-specific low-level statistics. This may support stronger zero-shot transfer.
% GTOT differs from the training data in acquisition conditions and modality-specific appearance statistics and exhibits comparatively limited target and scene diversity. This shift may limit Pre-AFA transferability, as its pre-encoder alignment can remain sensitive to input domain statistics. In contrast, Enc-CFA aligns target-conditioned template–search representations after encoder interaction, reducing reliance on modality-specific low-level cues alignment and supporting stronger zero-shot transfer.
% Unlike these RGB-T benchmarks, GTOT contains grayscale-thermal pairs and exhibits comparatively limited target and scene diversity. The resulting shift in visible-domain statistics reduces Pre-AFA transferability. Enc-CFA instead aligns same-target template-search representations after encoder interaction, making it less sensitive to low-level appearance cues. This target-centric alignment better suits GTOT's grayscale-T setting.
% Pre-AFA still achieves the best T$\rightarrow$RGB precision, but Enc-CFA provides the strongest overall GTOT performance, suggesting that encoder-level contrastive alignment is particularly effective when cross-modal correspondence is less disrupted by large appearance changes.
\subsubsection{Attribute-Based Performance on RGBT234.}
\label{sec:appendix_attribute_based}
\begin{figure}[!t]
    \centering
    \includegraphics[width=0.32\linewidth]
    {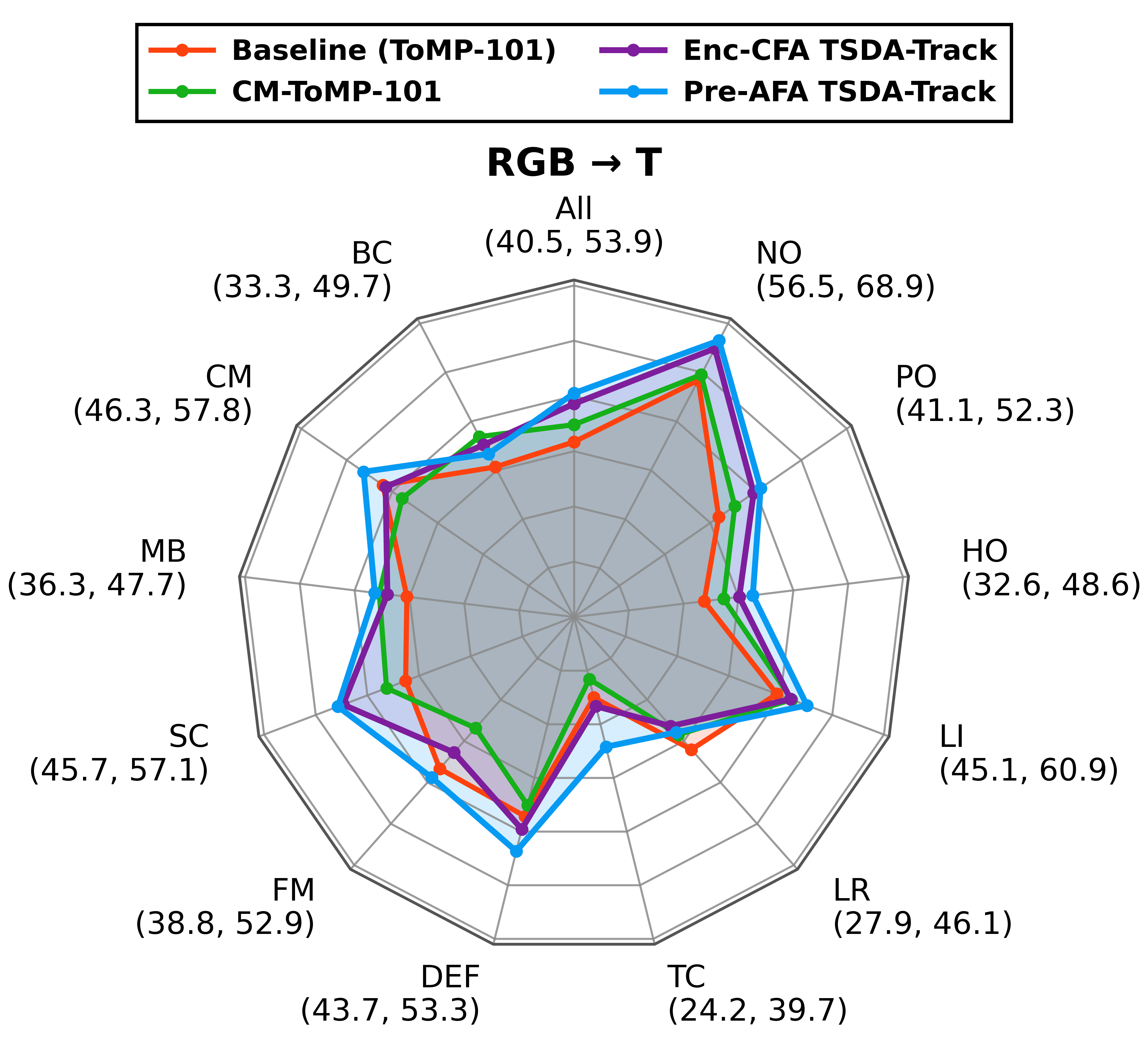}
    \hfill
    \includegraphics[width=0.32\linewidth]
    {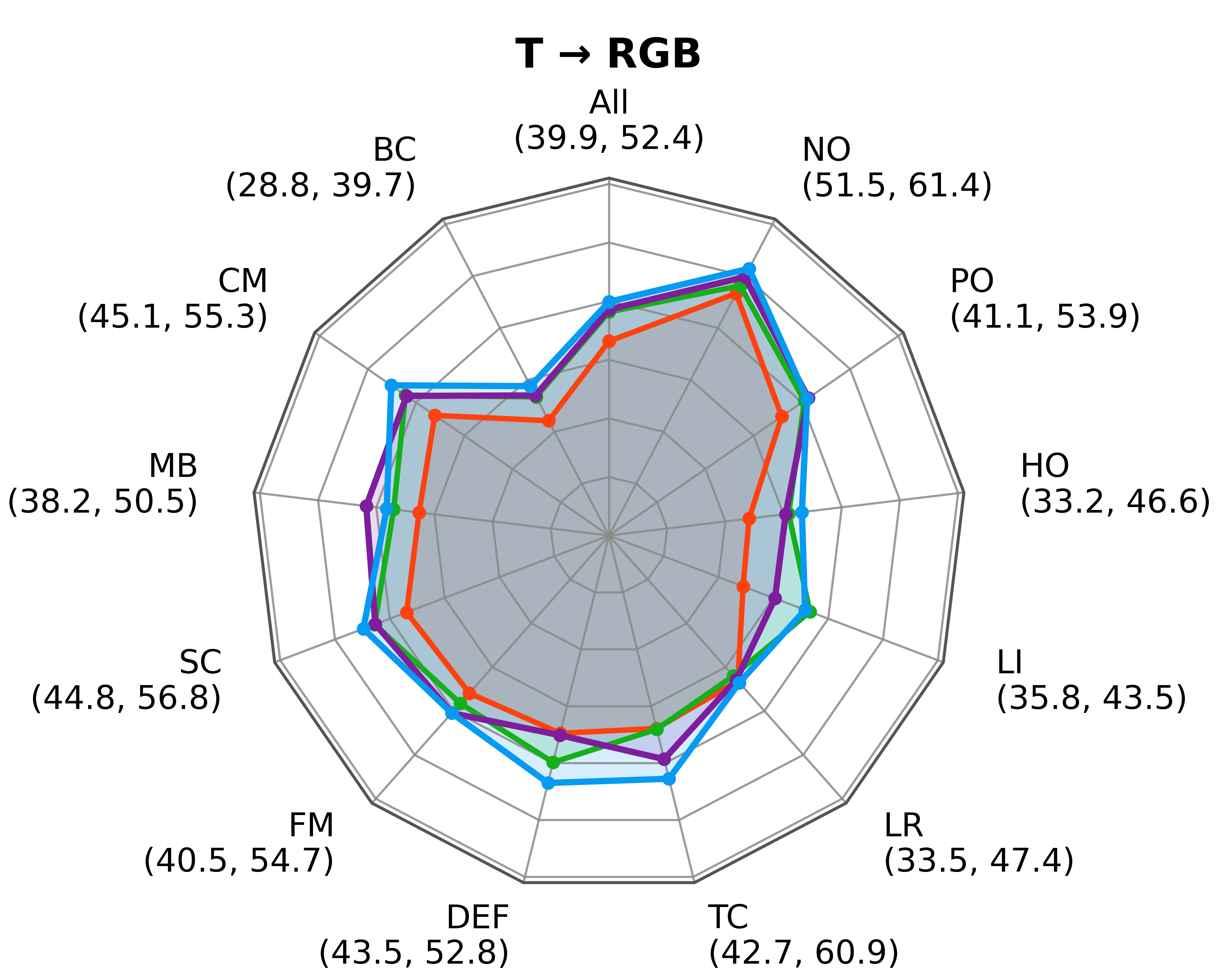}
    \hfill
    \includegraphics[width=0.32\linewidth]
    {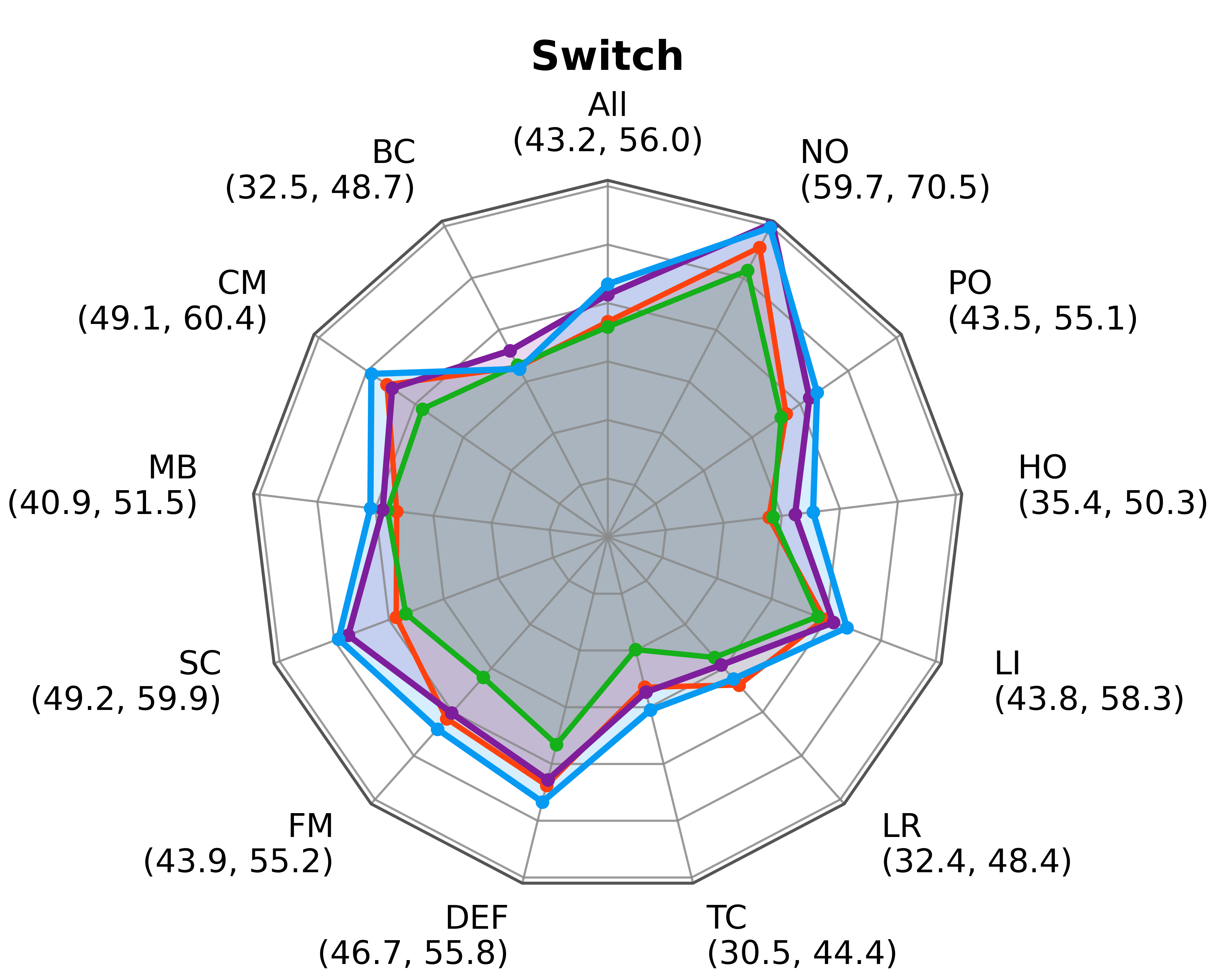}
    \caption{Attribute-based (SR$_{0.5}$, PR) comparison on RGBT234~\cite{li2019rgb} under the RGB$\rightarrow$T, T$\rightarrow$RGB, and Switch protocols.}
    \label{fig:attribute_rgbt234_sr50}
\end{figure}

To examine robustness under different tracking challenges, we evaluate the methods using the official sequence-level attributes of RGBT234~\cite{li2019rgb}. For each attribute, $SR_{0.5}$ and PR are averaged over all sequences based on the corresponding annotations. As shown in Figure~\ref{fig:attribute_rgbt234_sr50}, cross-modal training alone provides inconsistent gains over ToMP-101, confirming that exposure to mismatched template-search modalities does not sufficiently address their representation gap. Pre-AFA achieves the strongest overall attribute performance, with clear improvements under several challenging conditions, including occlusion, illumination variation, thermal
crossover, deformation, scale change, motion blur, and camera motion. This supports the benefit of suppressing modality-dependent appearance bias before template-search interaction. Enc-CFA also improves multiple
attributes by strengthening target-level correspondence after encoder interaction, although its gains on RGBT234 are less consistent than those of Pre-AFA. Performance remains comparatively limited under
low-resolution and background-clutter, indicating that feature alignment alone cannot fully resolve information loss and target-distractor ambiguity.

\subsubsection{Success and Precision Plots.}
\label{sec:appendix_success_precision}

\begin{figure}[!t]
    \centering

    % ==================== Success row ====================
    \begin{subfigure}[h]{\textwidth}
        \centering
        \includegraphics[width=0.32\linewidth]
        {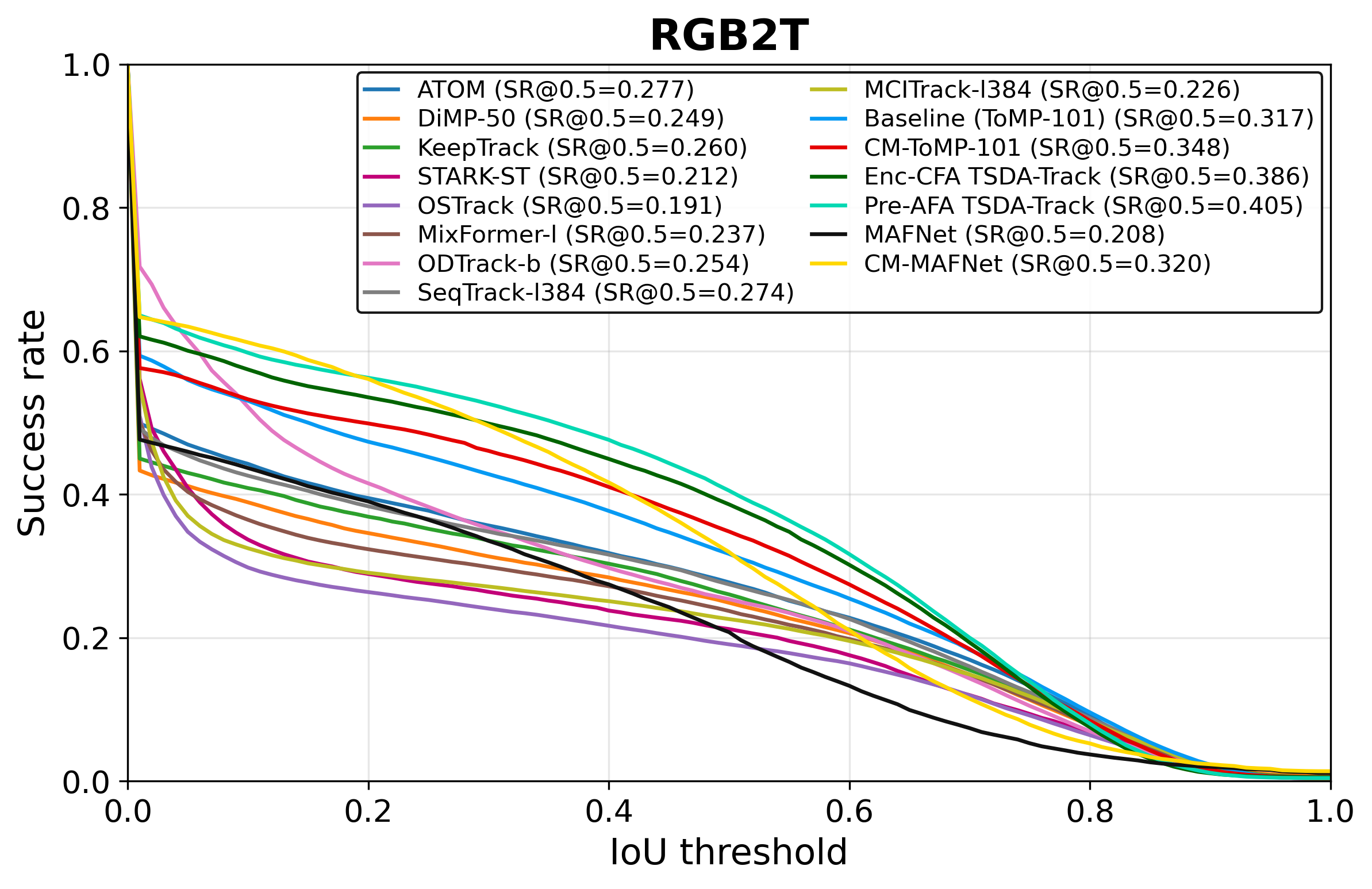}
        \hfill
        \includegraphics[width=0.32\linewidth]
        {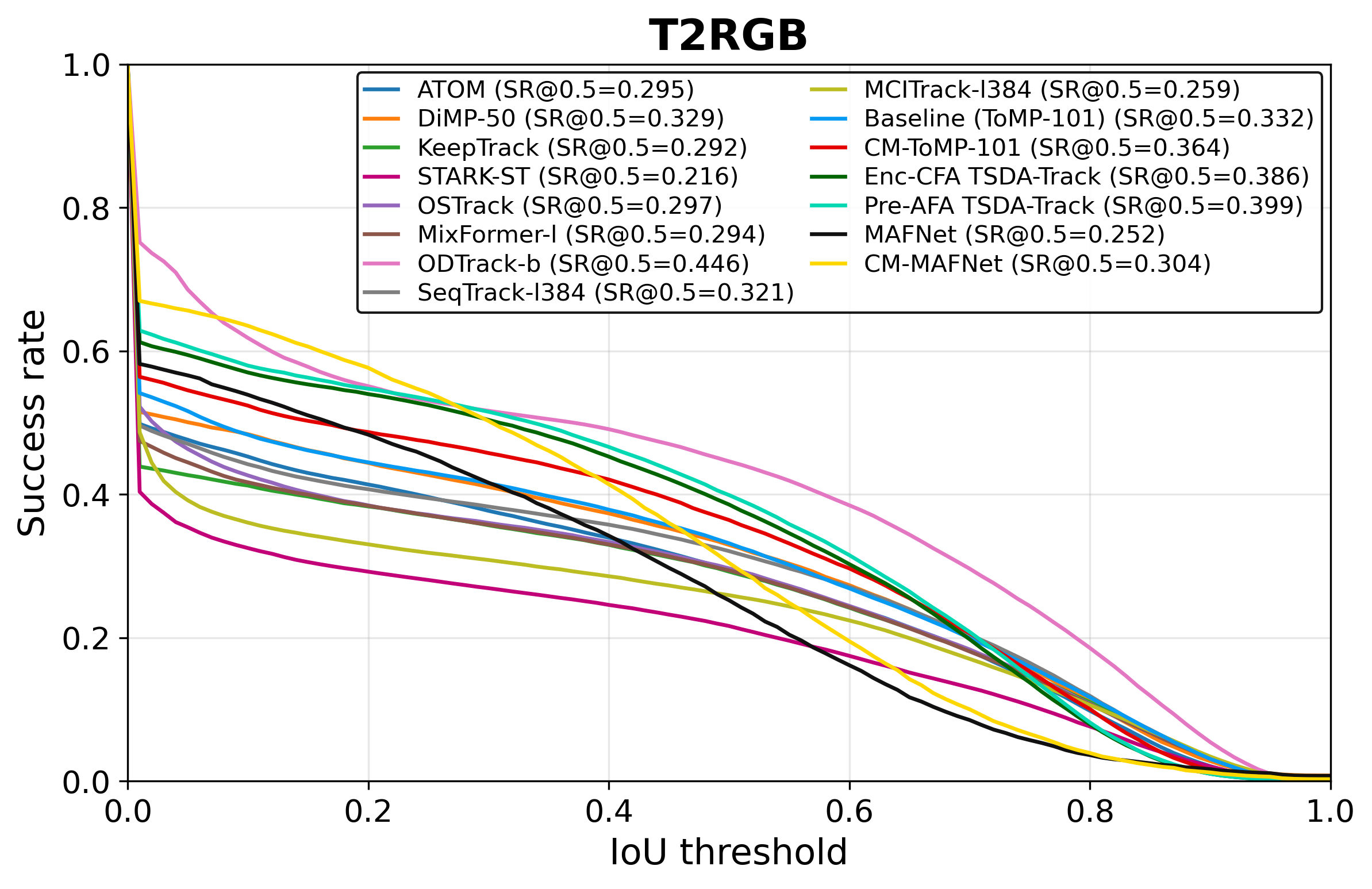}
        \hfill
        \includegraphics[width=0.32\linewidth]
        {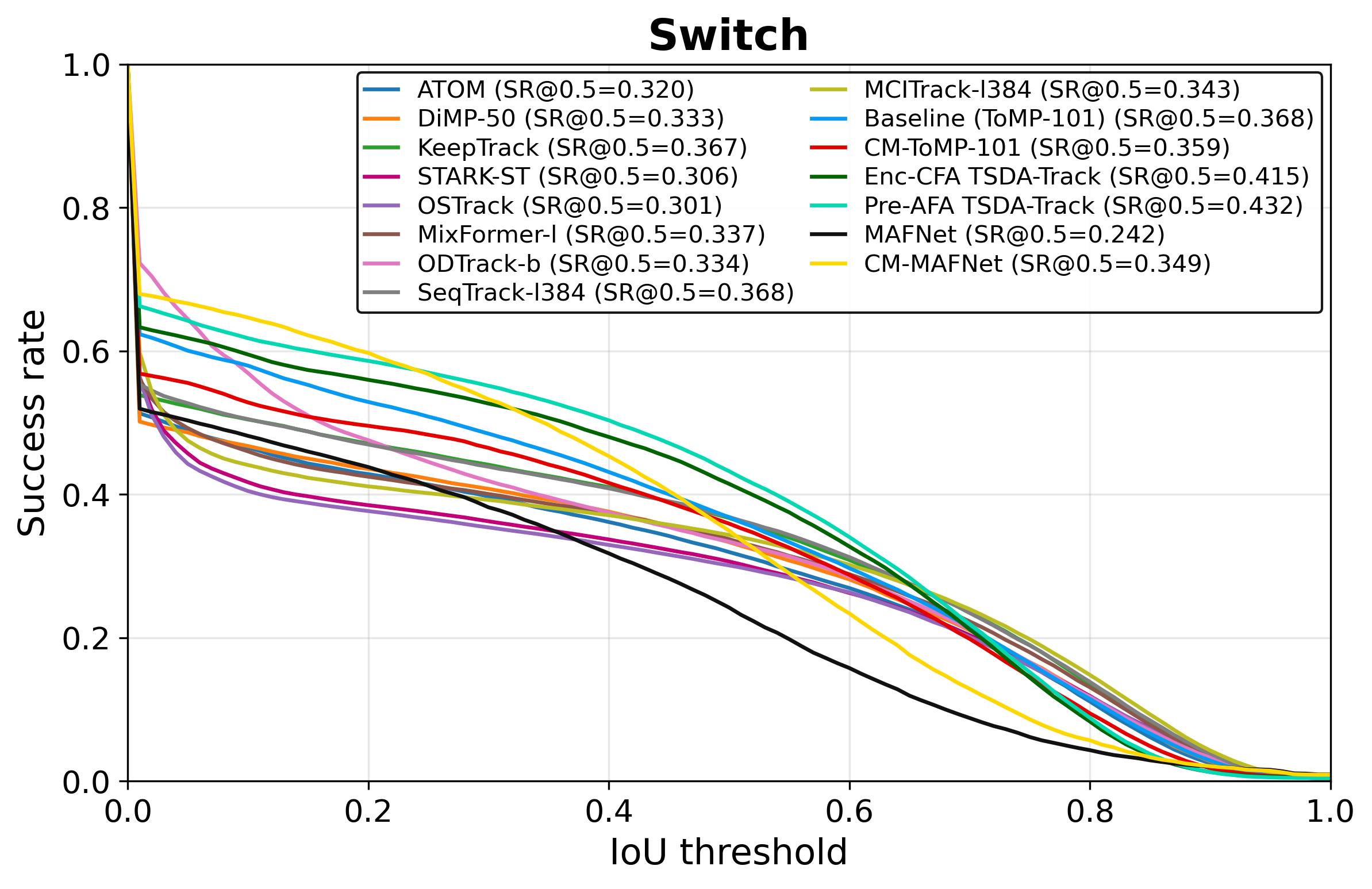}
        \caption{Success plots under the RGB$\rightarrow$T, T$\rightarrow$RGB, and Switch settings.}
        \label{fig:success_plots_all}
    \end{subfigure}

    \vspace{0.6em}

    % ==================== Precision row ====================
    \begin{subfigure}[h]{\textwidth}
        \centering
        \includegraphics[width=0.32\linewidth]
        {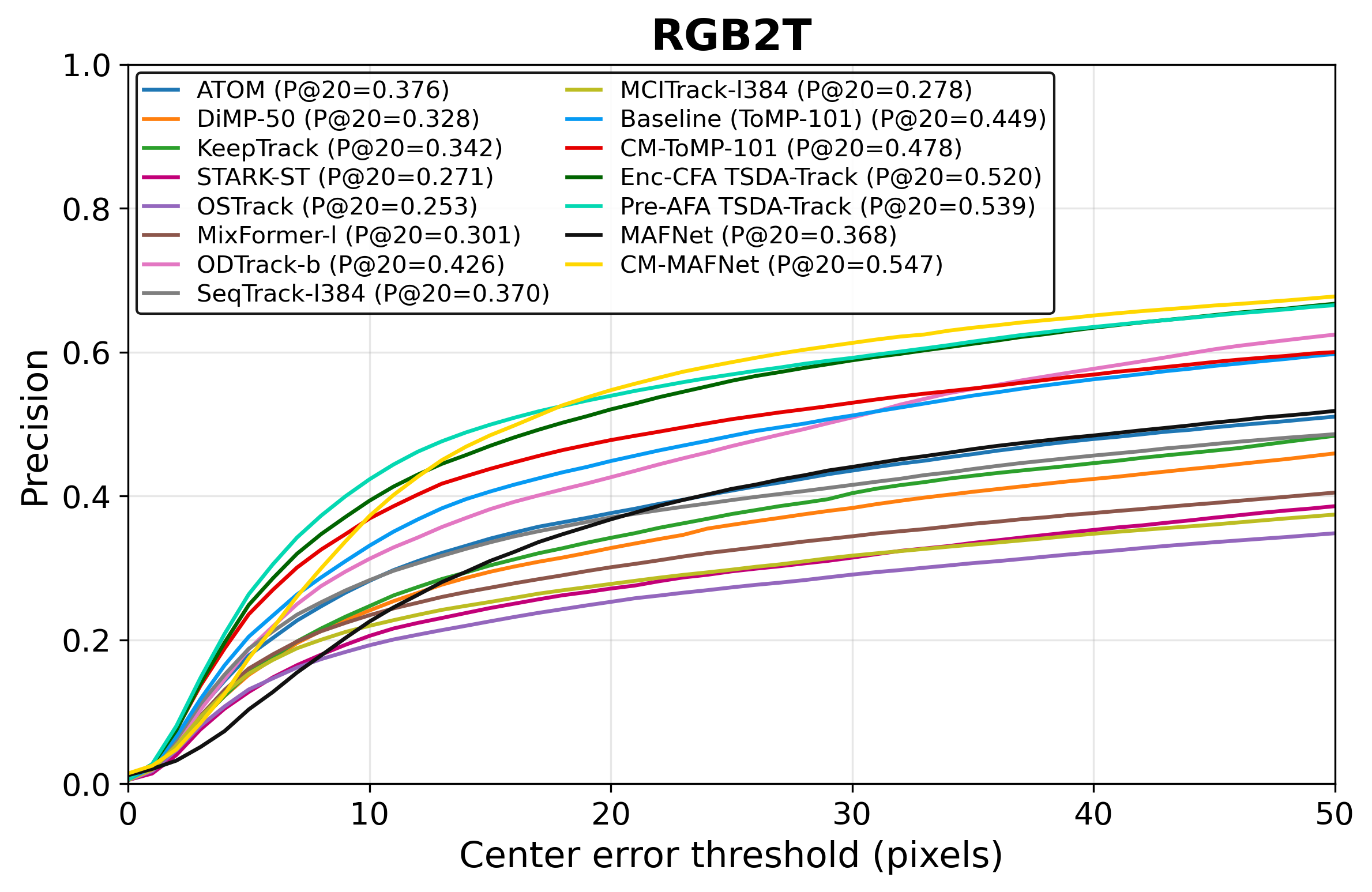}
        \hfill
        \includegraphics[width=0.32\linewidth]
        {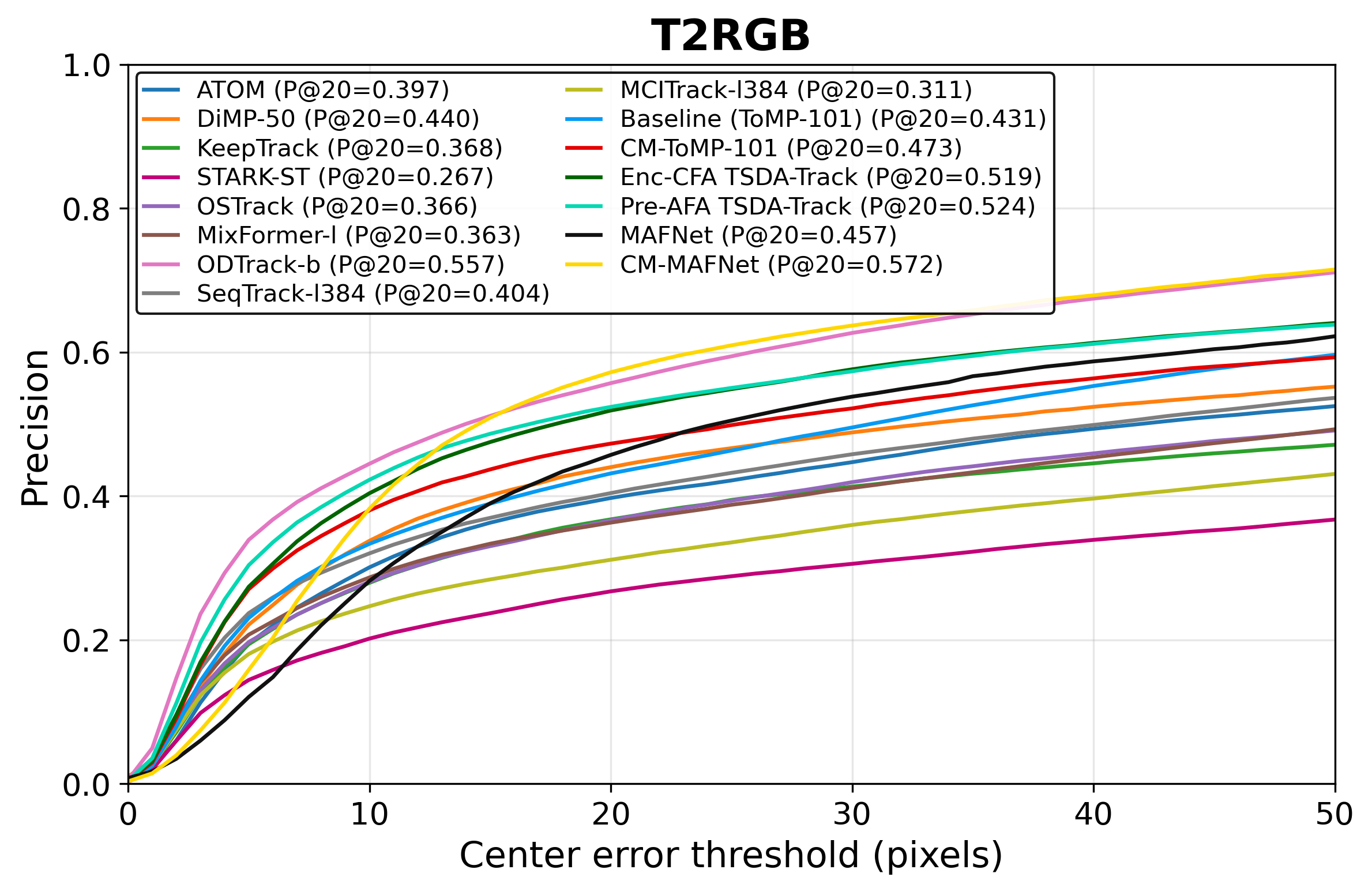}
        \hfill
        \includegraphics[width=0.32\linewidth]
        {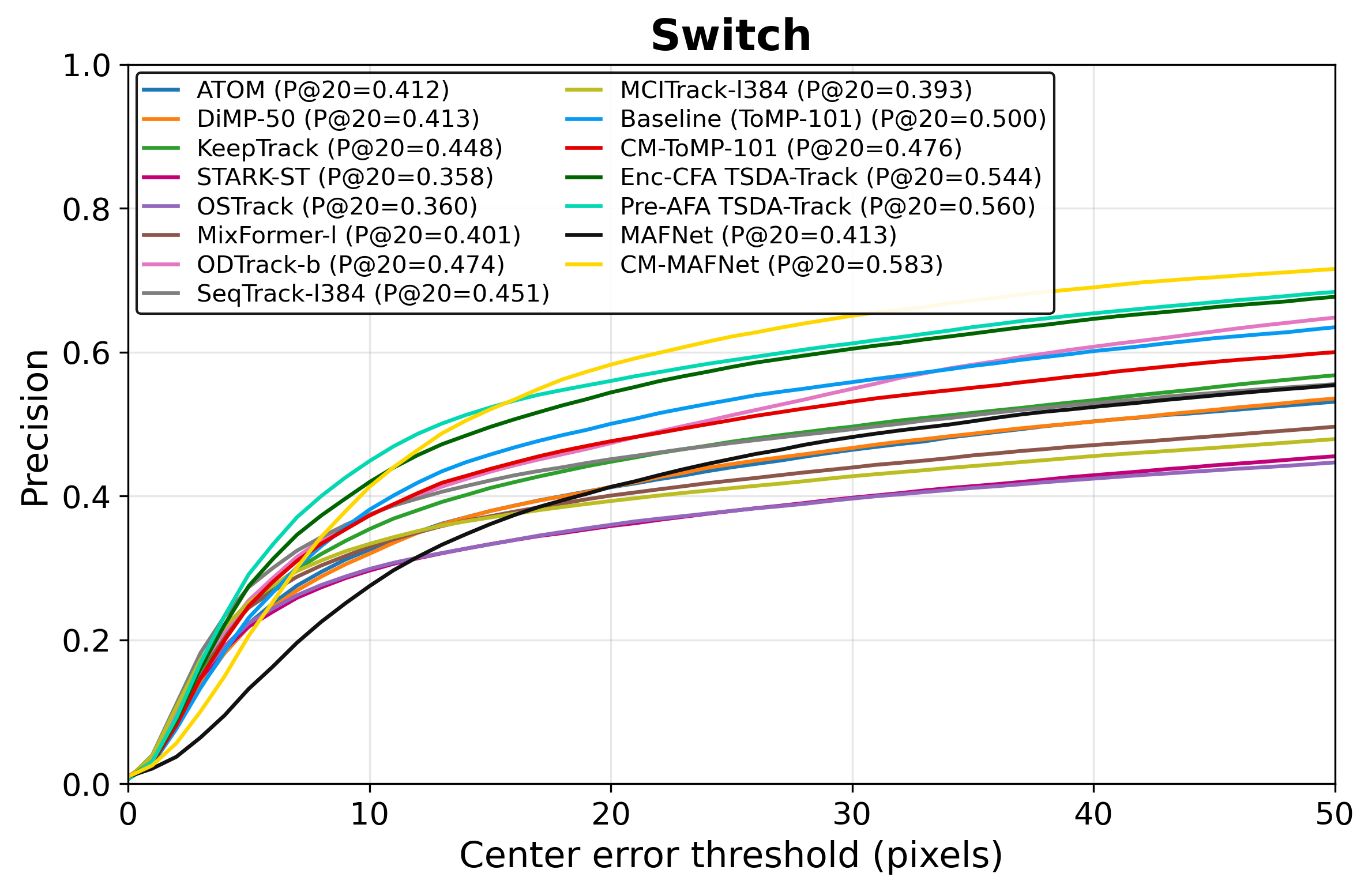}
        \caption{Precision plots under the RGB$\rightarrow$T, T$\rightarrow$RGB, and Switch settings.}
        \label{fig:precision_plots_all}
    \end{subfigure}

    \caption{Tracking performance on RGBT234~\cite{{li2019rgb}} under the RGB$\rightarrow$T, T$\rightarrow$RGB, and Switch evaluation protocols.}
    \label{fig:success_precision_all}
\end{figure}

Figure~\ref{fig:success_precision_all} presents the full success and precision curves on RGBT234~\cite{li2019rgb} under RGB$\rightarrow$T, T$\rightarrow$RGB, and Switch cross-modality protocols. Pre-AFA TSDA-Track shows the most consistent overall robustness, with particularly promising success rate curves under RGB$\rightarrow$T and Switch. Enc-CFA also consistently improves over ToMP-101 and CM-ToMP-101. ODTrack~\cite{zheng2024odtrack} achieves strong performance in T$\rightarrow$RGB with the highest SR$_{0.5}$ in this protocol, but this advantage is direction-specific, as its substantially weaker RGB$\rightarrow$T and Switch performance indicates limited consistency across configurations. MAFNet~\cite{liu2024cross} performs poorly under direct transfer from RGB-NIR to RGB-T. CM-MAFNet benefits from RGB-T cross-modal training and achieves high precision across all three protocols, but its SR remains substantially lower than that of the TSDA-Track variants indicating that improved target-center localization does not translate into equally reliable bounding-box overlap and scale estimation. Overall, the curves show that explicit feature alignment provides more consistent cross-modal robustness across protocols than either direction-specific gains or cross-modal training alone.

\subsubsection{Aerial Cross-Modal Tracking.}
\begin{figure}[!b]
    \centering
    %==================== (a) Success row ====================
    \begin{subfigure}[t]{\columnwidth}
        \centering
        \includegraphics[width=0.99\linewidth]{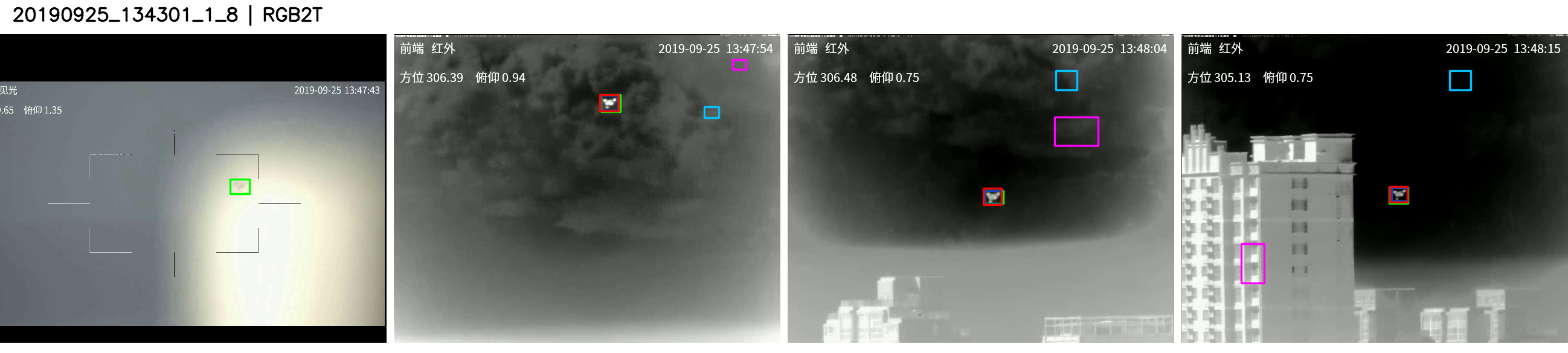}
        \label{}
    \end{subfigure}
    \begin{subfigure}[t]{\columnwidth}
        \centering        \includegraphics[width=0.99\linewidth]{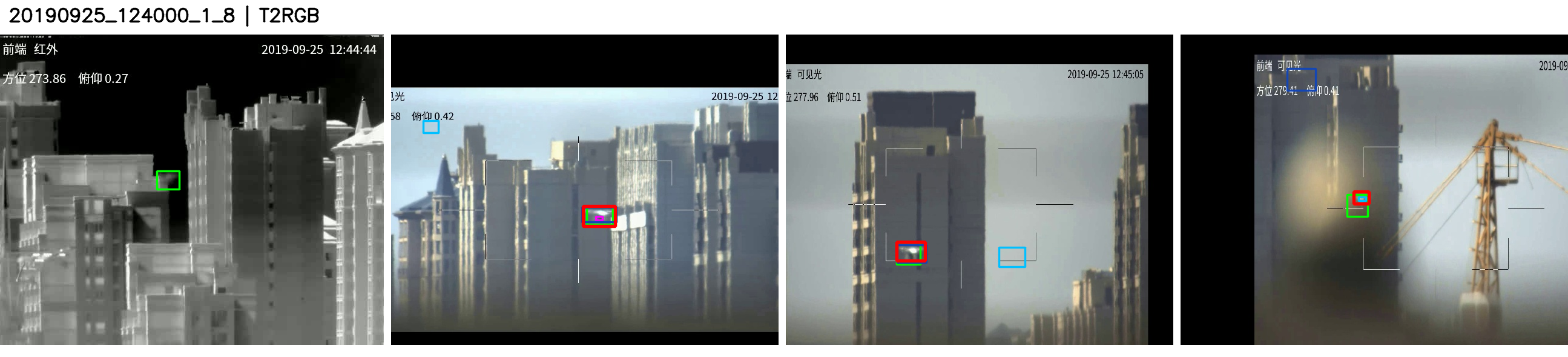}       
        \label{}
    \end{subfigure}
    \begin{subfigure}[t]{\columnwidth}
        \centering
        \includegraphics[width=0.99\linewidth]{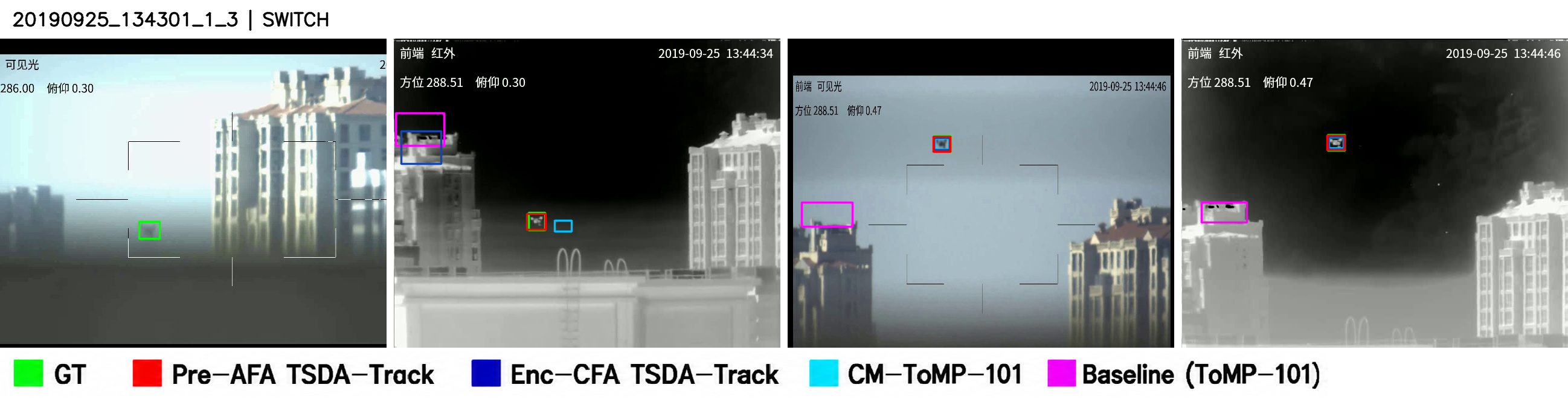}
        \label{}
    \end{subfigure}    
    \caption{Visual comparison on Anti-UAV-024~\cite{jiang2021anti} test split under RGB$\rightarrow$T, T$\rightarrow$RGB, and Switch mode.}
    \label{fig:Anti-UAV-024RGBT}
\end{figure}

We conduct a study on aerial RGB-T Anti-UAV-024~\cite{jiang2021anti}. Since its RGB and thermal videos have drastic spatial and resolution mismatch, we align the dataset using thermal frames as the reference and apply affine normalization preprocessing on RGB frames to reduce registration and scale mismatch. Models are trained on the training split and evaluated on the test split under the cross-modal protocols.
Table~\ref{tab:Anti-UAV-024RGBT} shows that Pre-AFA performs best under all protocols. Relative to CM-ToMP-101, it improves SR$_{0.5}$/PR by $6.5/6.6$, $5.2/4.9$, and $4.4/4.1$ points for RGB$\rightarrow$T, T$\rightarrow$RGB, and Switch, respectively. Enc-CFA consistently ranks second, while Pre-AFA+Enc-CFA remains inferior, indicating that they are not reliably additive as joint alignment may suppress useful tracking information. Qualitative results are shown in Figure~\ref{fig:Anti-UAV-024RGBT}.
% showing that Pre-AFA TSDA-Track more closely follows the ground-truth target extent, while Baseline and CM-Baseline exhibit greater localization drift or scale errors after cross-modal transitions.

\section{Limitations}
TSDA-Track improves cross-modal robustness through feature alignment but does not explicitly address geometric mismatch, which may limit bounding-box regression under pixel-level misalignment, scale or aspect-ratio variation, and clutter. Future work will explore joint feature geometry-aware alignment, develop large-scale benchmarks covering broader modality transitions, and extend TSDA-Track to additional modality pairs such as RGB-NIR, RGB-depth, and event data. Moreover, TSDA-Track is architecture-dependent rather than backbone-agnostic as its alignment modules are built on ToMP with explicit access to pre- and post-interaction stages. Extending it to one-stream trackers such as OSTrack and MixFormer requires alignment reformulation since template and search tokens interact from early stages.
% Moreover, TSDA-Track is built on ToMP with explicit clean access to pre- and post-interaction stages. Extending it to one-stream trackers such as OSTrack requires alignment reformulation since template and search tokens interact from early layers.
% TSDA-Track improves cross-modal performance through feature-level alignment, but residual geometric mismatch can still affect localization. Different sensors may introduce pixel-level misalignment and scale or aspect-ratio differences, leaving bounding-box regression challenging for small targets, scale changes, and cluttered scenes. Future work will explore joint feature and geometry-aware alignment for more accurate cross-modal localization, develop benchmarks with various modality transition protocols, and extend TSDA-Track to other modality pairs, including RGB-NIR, RGB-depth, and event-based tracking.

\begin{table}[!tb]
\centering
\caption{Cross-modal tracking performance on Anti-UAV-024~\cite{jiang2021anti} dataset. Best and second-best results are shown in bold and underline, respectively.
}
\label{tab:Anti-UAV-024RGBT}
\renewcommand{\arraystretch}{1.05}
\setlength{\tabcolsep}{2.2pt}
\begin{adjustbox}{max width=\columnwidth}
\begin{tabular}{@{}lcccccc@{}}
\toprule
\multirow{2}{*}{Method}
& \multicolumn{2}{c}{RGB$\rightarrow$T}
& \multicolumn{2}{c}{T$\rightarrow$RGB}
& \multicolumn{2}{c}{Switch} \\
\cmidrule(l{2pt}r{2pt}){2-3}
\cmidrule(l{2pt}r{2pt}){4-5}
\cmidrule(l{2pt}r{2pt}){6-7}
& SR$_{0.5}$ & PR
& SR$_{0.5}$ & PR
& SR$_{0.5}$ & PR \\
\midrule
ToMP-101~\cite{mayer2022transforming}
& 61.9 & 63.9
& 56.1 & 67.1
& 69.5 & 71.9 \\
CM-ToMP-101
& 69.7 & 71.9
& 77.8 & 79.8
& 75.2 & 77.9 \\
Enc-CFA TSDA-Track
& \underline{74.3} & \underline{78.1}
& \underline{81.5} & \underline{84.2}
& \underline{77.4} & \underline{81.2} \\
Pre-AFA TSDA-Track
& \textbf{76.2} & \textbf{78.5}
& \textbf{83.0} & \textbf{84.7}
& \textbf{79.6} & \textbf{82.0} \\
Pre-AFA + Enc-CFA TSDA-Track
& 72.0 & 75.7
& 79.5 & 83.5
& 76.2 & 79.8 \\
\bottomrule
\end{tabular}
\end{adjustbox}
\end{table}

\section{Conclusions}
This paper introduced TSDA-Track, a template-search domain adaptation framework for cross-modal tracking with two training-only alignment strategies. Pre-AFA suppresses modality-specific bias through adversarial alignment before transformer template-search interaction at encoder, while Enc-CFA strengthens target-level cross-modal correspondence through post-encoder contrastive alignment. Zero-shot results on RGBT234 and GTOT, together with evaluations on LasHeR and Anti-UAV-024, show that explicit alignment outperforms representative single-modal and prior cross-modal trackers across cross-modal switching protocols while retaining a shared inference pipeline without modality-specific routing. CM-ToMP-101 and CM-MAFNet further show that cross-modal training alone is insufficient for stable performance under modality discrepancy. Pre-AFA is more consistent under broader appearance variation, while Enc-CFA shows stronger transfer under shifts in acquisition conditions and modality-specific statistics, benefiting from its target-level semantic alignment. This highlights the importance of alignment objective and its integration stage in cross-modal tracking.
% Pre-AFA is more consistent under broader appearance variation, whereas Enc-CFA is more effective when low-level visible cues are limited, benefiting from its target-level semantic alignment. This highlights the importance of alignment objective and its integration stage in cross-modal tracking.
% This paper introduced TSDA-Track, a template-search domain adaptation framework for cross-modal tracking, and examined two training-only alignment strategies. Pre-AFA suppresses modality-specific bias through adversarial alignment before transformer template-search interaction, while Enc-CFA strengthens target-level cross-modal correspondence via contrastive alignment after encoder interaction. Zero-shot experiments on RGBT234 and GTOT, and evaluation on LasHer and spatially-aligned Anti-UAV-024 show that explicit feature alignment improves performance across multiple swiching protocols compared with repersentative single-modal and cross-modal trackers while preserving a shared inference pipeline without modality-specific routing. Pre-AFA performs better across datasets with broader RGB-thermal appearance and modality variation, whereas Enc-CFA is particularly effective when low-level visible cues are limited benefiting from benefit of target-level semantic alignment. These findings highlight the importance of alignment objective and injection stage in cross-modal tracking.

\bibliographystyle{unsrtnat}
\bibliography{references}

\begin{thebibliography}{42}
\providecommand{\natexlab}[1]{#1}
\providecommand{\url}[1]{\texttt{#1}}
\expandafter\ifx\csname urlstyle\endcsname\relax
  \providecommand{\doi}[1]{doi: #1}\else
  \providecommand{\doi}{doi: \begingroup \urlstyle{rm}\Url}\fi

\bibitem[Liu et~al.(2025)Liu, Li, Wang, Shen, and Tang]{liu2025prototype}
Lei Liu, Chenglong Li, Futian Wang, Longfeng Shen, and Jin Tang.
\newblock Prototype-based cross-modal object tracking.
\newblock \emph{Information Fusion}, 118:\penalty0 102941, 2025.

\bibitem[Liu et~al.(2024{\natexlab{a}})Liu, Zhang, Li, Li, and Tang]{liu2024cross}
Lei Liu, Mengya Zhang, Cheng Li, Chenglong Li, and Jin Tang.
\newblock Cross-modal object tracking via modality-aware fusion network and a large-scale dataset.
\newblock \emph{IEEE Transactions on Neural Networks and Learning Systems}, 36\penalty0 (4):\penalty0 6981--6994, 2024{\natexlab{a}}.

\bibitem[Li et~al.(2022)Li, Zhu, Liu, Si, Fan, and Zhai]{li2022cross}
Chenglong Li, Tianhao Zhu, Lei Liu, Xiaonan Si, Zilin Fan, and Sulan Zhai.
\newblock Cross-modal object tracking: Modality-aware representations and a unified benchmark.
\newblock In \emph{Proceedings of the AAAI Conference on Artificial Intelligence}, volume~36, pages 1289--1296, 2022.

\bibitem[Xu et~al.(2025)Xu, Hou, Ren, Zhou, Wu, and Cao]{xu2025switrack}
Boyue Xu, Ruichao Hou, Tongwei Ren, Dongming Zhou, Gangshan Wu, and Jinde Cao.
\newblock Switrack: Tri-state switch for cross-modal object tracking.
\newblock \emph{arXiv preprint arXiv:2511.16227}, 2025.

\bibitem[Ganin et~al.(2016)Ganin, Ustinova, Ajakan, Germain, Larochelle, Laviolette, March, and Lempitsky]{ganin2016domain}
Yaroslav Ganin, Evgeniya Ustinova, Hana Ajakan, Pascal Germain, Hugo Larochelle, Fran{\c{c}}ois Laviolette, Mario March, and Victor Lempitsky.
\newblock Domain-adversarial training of neural networks.
\newblock \emph{Journal of machine learning research}, 17\penalty0 (59):\penalty0 1--35, 2016.

\bibitem[Oord et~al.(2018)Oord, Li, and Vinyals]{oord2018representation}
Aaron van~den Oord, Yazhe Li, and Oriol Vinyals.
\newblock Representation learning with contrastive predictive coding.
\newblock \emph{arXiv preprint arXiv:1807.03748}, 2018.

\bibitem[Pang et~al.(2021)Pang, Qiu, Li, Chen, Li, Darrell, and Yu]{pang2021quasi}
Jiangmiao Pang, Linlu Qiu, Xia Li, Haofeng Chen, Qi~Li, Trevor Darrell, and Fisher Yu.
\newblock Quasi-dense similarity learning for multiple object tracking.
\newblock In \emph{2021 IEEE/CVF Conference on Computer Vision and Pattern Recognition (CVPR)}, pages 164--173, 2021.
\newblock \doi{10.1109/CVPR46437.2021.00023}.

\bibitem[Meibodi et~al.(2025)Meibodi, Alijani, and Najjaran]{meibodi2025deep}
Fereshteh~Aghaee Meibodi, Shadi Alijani, and Homayoun Najjaran.
\newblock A deep dive into generic object tracking: A survey.
\newblock \emph{arXiv preprint arXiv:2507.23251}, 2025.

\bibitem[Danelljan et~al.(2019)Danelljan, Bhat, Khan, and Felsberg]{danelljan2019atom}
Martin Danelljan, Goutam Bhat, Fahad~Shahbaz Khan, and Michael Felsberg.
\newblock Atom: Accurate tracking by overlap maximization.
\newblock In \emph{Proceedings of the IEEE/CVF Conference on Computer Vision and Pattern Recognition (CVPR)}, June 2019.

\bibitem[Bhat et~al.(2019)Bhat, Danelljan, Gool, and Timofte]{bhat2019learning}
Goutam Bhat, Martin Danelljan, Luc~Van Gool, and Radu Timofte.
\newblock Learning discriminative model prediction for tracking.
\newblock In \emph{Proceedings of the IEEE/CVF International Conference on Computer Vision (ICCV)}, October 2019.

\bibitem[Yu et~al.(2020)Yu, Xiong, Huang, and Scott]{yu2020deformable}
Yuechen Yu, Yilei Xiong, Weilin Huang, and Matthew~R. Scott.
\newblock Deformable siamese attention networks for visual object tracking.
\newblock In \emph{Proceedings of the IEEE/CVF Conference on Computer Vision and Pattern Recognition (CVPR)}, June 2020.

\bibitem[Liu et~al.(2024{\natexlab{b}})Liu, Wang, Ma, Su, and Yang]{liu2024siamdmu}
Jing Liu, Han Wang, Chao Ma, Yuting Su, and Xiaokang Yang.
\newblock Siamdmu: Siamese dual mask update network for visual object tracking.
\newblock \emph{IEEE Transactions on Emerging Topics in Computational Intelligence}, 8\penalty0 (2):\penalty0 1656--1669, 2024{\natexlab{b}}.
\newblock \doi{10.1109/TETCI.2024.3353674}.

\bibitem[Chen et~al.(2021)Chen, Yan, Zhu, Wang, Yang, and Lu]{chen2021transformer}
Xin Chen, Bin Yan, Jiawen Zhu, Dong Wang, Xiaoyun Yang, and Huchuan Lu.
\newblock Transformer tracking.
\newblock In \emph{Proceedings of the IEEE/CVF conference on computer vision and pattern recognition}, pages 8126--8135, 2021.

\bibitem[Yan et~al.(2021)Yan, Peng, Fu, Wang, and Lu]{yan2021learning}
Bin Yan, Houwen Peng, Jianlong Fu, Dong Wang, and Huchuan Lu.
\newblock Learning spatio-temporal transformer for visual tracking.
\newblock In \emph{Proceedings of the IEEE/CVF international conference on computer vision}, pages 10448--10457, 2021.

\bibitem[Mayer et~al.(2022)Mayer, Danelljan, Bhat, Paul, Paudel, Yu, and Van~Gool]{mayer2022transforming}
Christoph Mayer, Martin Danelljan, Goutam Bhat, Matthieu Paul, Danda~Pani Paudel, Fisher Yu, and Luc Van~Gool.
\newblock Transforming model prediction for tracking.
\newblock In \emph{Proceedings of the IEEE/CVF conference on computer vision and pattern recognition}, pages 8731--8740, 2022.

\bibitem[Mayer et~al.(2024)Mayer, Danelljan, Yang, Ferrari, Van~Gool, and Kuznetsova]{mayer2024beyond}
Christoph Mayer, Martin Danelljan, Ming-Hsuan Yang, Vittorio Ferrari, Luc Van~Gool, and Alina Kuznetsova.
\newblock Beyond sot: Tracking multiple generic objects at once.
\newblock In \emph{Proceedings of the IEEE/CVF Winter Conference on Applications of Computer Vision}, pages 6826--6836, 2024.

\bibitem[Ye et~al.(2022{\natexlab{a}})Ye, Chang, Ma, Shan, and Chen]{ye2022joint}
Botao Ye, Hong Chang, Bingpeng Ma, Shiguang Shan, and Xilin Chen.
\newblock Joint feature learning and relation modeling for tracking: A one-stream framework.
\newblock In Shai Avidan, Gabriel Brostow, Moustapha Ciss{\'e}, Giovanni~Maria Farinella, and Tal Hassner, editors, \emph{Computer Vision -- ECCV 2022}, pages 341--357, Cham, 2022{\natexlab{a}}. Springer Nature Switzerland.
\newblock ISBN 978-3-031-20047-2.

\bibitem[Cui et~al.(2024)Cui, Jiang, Wu, and Wang]{cui2022mixformer}
Yutao Cui, Cheng Jiang, Gangshan Wu, and Limin Wang.
\newblock Mixformer: End-to-end tracking with iterative mixed attention.
\newblock \emph{IEEE Transactions on Pattern Analysis and Machine Intelligence}, 46\penalty0 (6):\penalty0 4129--4146, 2024.
\newblock \doi{10.1109/TPAMI.2024.3349519}.

\bibitem[Wei et~al.(2023)Wei, Bai, Zheng, Shi, and Gong]{wei2023autoregressive}
Xing Wei, Yifan Bai, Yongchao Zheng, Dahu Shi, and Yihong Gong.
\newblock Autoregressive visual tracking.
\newblock In \emph{Proceedings of the IEEE/CVF Conference on Computer Vision and Pattern Recognition (CVPR)}, pages 9697--9706, June 2023.

\bibitem[Chen et~al.(2023)Chen, Peng, Wang, Lu, and Hu]{chen2023seqtrack}
Xin Chen, Houwen Peng, Dong Wang, Huchuan Lu, and Han Hu.
\newblock Seqtrack: Sequence to sequence learning for visual object tracking.
\newblock In \emph{Proceedings of the IEEE/CVF Conference on Computer Vision and Pattern Recognition (CVPR)}, pages 14572--14581, June 2023.

\bibitem[Zheng et~al.(2024)Zheng, Zhong, Liang, Mo, Zhang, and Li]{zheng2024odtrack}
Yaozong Zheng, Bineng Zhong, Qihua Liang, Zhiyi Mo, Shengping Zhang, and Xianxian Li.
\newblock Odtrack: Online dense temporal token learning for visual tracking.
\newblock In \emph{Proceedings of the AAAI conference on artificial intelligence}, volume~38, pages 7588--7596, 2024.

\bibitem[Kang et~al.(2025)Kang, Chen, Lai, Liu, Liu, and Wang]{kang2025exploring}
Ben Kang, Xin Chen, Simiao Lai, Yang Liu, Yi~Liu, and Dong Wang.
\newblock Exploring enhanced contextual information for video-level object tracking.
\newblock In \emph{Proceedings of the AAAI conference on artificial intelligence}, volume~39, pages 4194--4202, 2025.

\bibitem[Cai et~al.(2023)Cai, Liu, Tang, and Wu]{cai2023robust}
Yidong Cai, Jie Liu, Jie Tang, and Gangshan Wu.
\newblock Robust object modeling for visual tracking.
\newblock In \emph{Proceedings of the IEEE/CVF International Conference on Computer Vision (ICCV)}, pages 9589--9600, October 2023.

\bibitem[Xiao et~al.(2022)Xiao, Yang, Li, Liu, and Tang]{xiao2022attribute}
Yun Xiao, Mengmeng Yang, Chenglong Li, Lei Liu, and Jin Tang.
\newblock Attribute-based progressive fusion network for rgbt tracking.
\newblock In \emph{Proceedings of the AAAI conference on artificial intelligence}, volume~36, pages 2831--2838, 2022.

\bibitem[Hou et~al.(2023)Hou, Xu, Ren, and Wu]{hou2023mtnet}
Ruichao Hou, Boyue Xu, Tongwei Ren, and Gangshan Wu.
\newblock Mtnet: Learning modality-aware representation with transformer for rgbt tracking.
\newblock In \emph{2023 IEEE International Conference on Multimedia and Expo (ICME)}, pages 1163--1168. IEEE, 2023.

\bibitem[Yang et~al.(2022)Yang, Li, Zheng, Leonardis, and Song]{yang2022prompting}
Jinyu Yang, Zhe Li, Feng Zheng, Ales Leonardis, and Jingkuan Song.
\newblock Prompting for multi-modal tracking.
\newblock In \emph{Proceedings of the 30th ACM International Conference on Multimedia}, MM '22, page 3492–3500, New York, NY, USA, 2022. Association for Computing Machinery.
\newblock ISBN 9781450392037.
\newblock \doi{10.1145/3503161.3547851}.
\newblock URL \url{https://doi.org/10.1145/3503161.3547851}.

\bibitem[Zhu et~al.(2023)Zhu, Lai, Chen, Wang, and Lu]{zhu2023visual}
Jiawen Zhu, Simiao Lai, Xin Chen, Dong Wang, and Huchuan Lu.
\newblock Visual prompt multi-modal tracking.
\newblock In \emph{Proceedings of the IEEE/CVF conference on computer vision and pattern recognition}, pages 9516--9526, 2023.

\bibitem[Hou et~al.(2024)Hou, Xing, Qian, Guo, Xin, Chen, Tang, Wang, Jiang, Liu, et~al.]{hou2024sdstrack}
Xiaojun Hou, Jiazheng Xing, Yijie Qian, Yaowei Guo, Shuo Xin, Junhao Chen, Kai Tang, Mengmeng Wang, Zhengkai Jiang, Liang Liu, et~al.
\newblock Sdstrack: Self-distillation symmetric adapter learning for multi-modal visual object tracking.
\newblock In \emph{Proceedings of the IEEE/CVF conference on computer vision and pattern recognition}, pages 26551--26561, 2024.

\bibitem[Ye et~al.(2022{\natexlab{b}})Ye, Fu, Zheng, Paudel, and Chen]{ye2022unsupervised}
Junjie Ye, Changhong Fu, Guangze Zheng, Danda~Pani Paudel, and Guang Chen.
\newblock Unsupervised domain adaptation for nighttime aerial tracking.
\newblock In \emph{Proceedings of the IEEE/CVF conference on computer vision and pattern recognition}, pages 8896--8905, 2022{\natexlab{b}}.

\bibitem[Wei et~al.(2024)Wei, Fu, Wang, Guo, Zhao, and Fang]{wei2024unsupervised}
Haoran Wei, Yanyun Fu, Deyong Wang, Rui Guo, Xueyi Zhao, and Jian Fang.
\newblock Unsupervised nighttime object tracking based on transformer and domain adaptation fusion network.
\newblock \emph{IEEE Access}, 12:\penalty0 130896--130913, 2024.

\bibitem[Fu et~al.(2024{\natexlab{a}})Fu, Wang, Yao, Zheng, Zuo, and Pan]{fu2024prompt}
Changhong Fu, Yiheng Wang, Liangliang Yao, Guangze Zheng, Haobo Zuo, and Jia Pan.
\newblock Prompt-driven temporal domain adaptation for nighttime uav tracking.
\newblock In \emph{2024 IEEE/RSJ International Conference on Intelligent Robots and Systems (IROS)}, pages 9706--9713. IEEE, 2024{\natexlab{a}}.

\bibitem[Fu et~al.(2024{\natexlab{b}})Fu, Yao, Zuo, Zheng, and Pan]{fu2023sam}
Changhong Fu, Liangliang Yao, Haobo Zuo, Guangze Zheng, and Jia Pan.
\newblock Sam-da: Uav tracks anything at night with sam-powered domain adaptation.
\newblock In \emph{2024 International Conference on Advanced Robotics and Mechatronics (ICARM)}, pages 31--38, 2024{\natexlab{b}}.
\newblock \doi{10.1109/ICARM62033.2024.10715901}.

\bibitem[Li et~al.(2024)Li, Tan, Liu, Yuan, Li, and Liu]{li2024progressive}
Qiao Li, Kanlun Tan, Qiao Liu, Di~Yuan, Xin Li, and Yunpeng Liu.
\newblock Progressive domain adaptation for thermal infrared object tracking.
\newblock \emph{arXiv preprint arXiv:2407.19430}, 2024.

\bibitem[Li et~al.(2025)Li, Tan, Liu, Yuan, Li, and Liu]{li2025efficient}
Qiao Li, Kanlun Tan, Qiao Liu, Di~Yuan, Xin Li, and Yunpeng Liu.
\newblock Efficient hierarchical domain adaptive thermal infrared tracking.
\newblock In \emph{ICASSP 2025-2025 IEEE International Conference on Acoustics, Speech and Signal Processing (ICASSP)}, pages 1--5. IEEE, 2025.

\bibitem[Wu et~al.(2024)Wu, Jiao, Liu, Liu, Yang, and Li]{wu2024domain}
Yinan Wu, Licheng Jiao, Xu~Liu, Fang Liu, Shuyuan Yang, and Lingling Li.
\newblock Domain adaptation-aware transformer for hyperspectral object tracking.
\newblock \emph{IEEE Transactions on Circuits and Systems for Video Technology}, 34\penalty0 (9):\penalty0 8041--8052, 2024.

\bibitem[Ganin and Lempitsky(2015)]{ganin2015unsupervised}
Yaroslav Ganin and Victor Lempitsky.
\newblock Unsupervised domain adaptation by backpropagation.
\newblock In \emph{International conference on machine learning}, pages 1180--1189. PMLR, 2015.

\bibitem[Li et~al.(2021)Li, Xue, Jia, Qu, Luo, Tang, and Sun]{li2021lasher}
Chenglong Li, Wanlin Xue, Yaqing Jia, Zhichen Qu, Bin Luo, Jin Tang, and Dengdi Sun.
\newblock Lasher: A large-scale high-diversity benchmark for rgbt tracking.
\newblock \emph{IEEE Transactions on Image Processing}, 31:\penalty0 392--404, 2021.

\bibitem[Loshchilov and Hutter(2017)]{loshchilov2017decoupled}
Ilya Loshchilov and Frank Hutter.
\newblock Decoupled weight decay regularization.
\newblock \emph{arXiv preprint arXiv:1711.05101}, 2017.

\bibitem[Li et~al.(2019)Li, Liang, Lu, Zhao, and Tang]{li2019rgb}
Chenglong Li, Xinyan Liang, Yijuan Lu, Nan Zhao, and Jin Tang.
\newblock Rgb-t object tracking: Benchmark and baseline.
\newblock \emph{Pattern Recognition}, 96:\penalty0 106977, 2019.

\bibitem[Li et~al.(2016)Li, Cheng, Hu, Liu, Tang, and Lin]{li2016learning}
Chenglong Li, Hui Cheng, Shiyi Hu, Xiaobai Liu, Jin Tang, and Liang Lin.
\newblock Learning collaborative sparse representation for grayscale-thermal tracking.
\newblock \emph{IEEE Transactions on Image Processing}, 25\penalty0 (12):\penalty0 5743--5756, 2016.

\bibitem[Mayer et~al.(2021)Mayer, Danelljan, Paudel, and Van~Gool]{mayer2021learning}
Christoph Mayer, Martin Danelljan, Danda~Pani Paudel, and Luc Van~Gool.
\newblock Learning target candidate association to keep track of what not to track.
\newblock In \emph{Proceedings of the IEEE/CVF International Conference on Computer Vision (ICCV)}, pages 13444--13454, October 2021.

\bibitem[Jiang et~al.(2021)Jiang, Wang, Peng, Yu, Wang, Xing, Li, Guo, Ye, Jiao, et~al.]{jiang2021anti}
Nan Jiang, Kuiran Wang, Xiaoke Peng, Xuehui Yu, Qiang Wang, Junliang Xing, Guorong Li, Guodong Guo, Qixiang Ye, Jianbin Jiao, et~al.
\newblock Anti-uav: A large-scale benchmark for vision-based uav tracking.
\newblock \emph{IEEE Transactions on Multimedia}, 25:\penalty0 486--500, 2021.

\end{thebibliography}

\end{document}